\pdfoutput=1

\documentclass[11pt]{article}

\usepackage{ACL2023}

\usepackage{times}
\usepackage{latexsym}

\usepackage[T1]{fontenc}

\usepackage[utf8]{inputenc}

\usepackage{microtype}

\usepackage{inconsolata}

\usepackage{graphicx} 
\usepackage{makecell}
\usepackage{amsmath}
\usepackage{amsfonts}

\usepackage{multirow}
\usepackage{subfigure}

\usepackage{algorithm}
\usepackage{algorithmic}

\usepackage{afterpage}
\usepackage{lipsum}
\usepackage{float}
\usepackage{multicol} 
\usepackage{xspace}

\title{Are You Copying My Model? Protecting the Copyright of Large Language Models for EaaS via Backdoor Watermark}

\author{
  Wenjun Peng$^{1*}$, Jingwei Yi$^{1*}$, Fangzhao Wu$^{2\dag}$, Shangxi Wu$^3$, Bin Zhu$^2$, Lingjuan Lyu$^4$, \\ \textbf{Binxing Jiao}$^5$, \textbf{Tong Xu}$^{1\dag}$, \textbf{Guangzhong Sun}$^1$, \textbf{Xing Xie}$^2$ \\
  $^1$University of Science and Technology of China
  $^2$Microsoft Research Asia \\
  $^3$Beijing Jiaotong University
  $^4$Sony AI 
  $^5$Microsoft STC Asia \\
  {\tt \{pengwj,yjw1029\}@mail.ustc.edu.cn}
  {\tt wufangzhao@gmail.com} \\
  {\tt wushangxi@bjtu.edu.cn} 
  {\tt \{binzhu,binxjia,xingx\}@microsoft.com} \\
  {\tt lingjuan.lv@sony.com } {\tt \{tongxu,gzsun\}@ustc.edu.cn}
}

\newtheorem{define}{Definition}

\newtheorem{prop}{Proportion}
\newcommand{\method}{EmbMarker\xspace}

\begin{document}
\maketitle

\def\thefootnote{*}\footnotetext{Indicates equal contribution.}\def\thefootnote{\arabic{footnote}}

\def\thefootnote{\dag}\footnotetext{Corresponding authors.}\def\thefootnote{\arabic{footnote}}

\begin{abstract}
Large language models (LLMs) have demonstrated powerful capabilities in both text understanding and generation. 
Companies have begun to offer Embedding as a Service (EaaS) based on these LLMs, which can benefit various natural language processing (NLP) tasks for customers. 
However, previous studies have shown that EaaS is vulnerable to model extraction attacks, which can cause significant losses for the owners of LLMs, as training these models is extremely expensive. 
To protect the copyright of LLMs for EaaS, we propose an Embedding Watermark method called \method that implants backdoors on embeddings. 
Our method selects a group of moderate-frequency words from a general text corpus to form a trigger set, then selects a target embedding as the watermark, and inserts it into the embeddings of texts containing trigger words as the backdoor. 
The weight of insertion is proportional to the number of trigger words included in the text. 
This allows the watermark backdoor to be effectively transferred to EaaS-stealer's model for copyright verification while minimizing the adverse impact on the original embeddings' utility. 
Our extensive experiments on various datasets show that our method can effectively protect the copyright of EaaS models without compromising service quality.
Our code is available at \url{https://github.com/yjw1029/EmbMarker}.
\end{abstract}
\section{Introduction}

Large language models (LLMs) such as GPT-3~\cite{brown2020language} and LLAMA~\cite{touvron2023llama} have demonstrated exceptional abilities in natural language understanding and generation. 
As a result, the owners of these LLMs have started offering Embedding as a Service (EaaS) to assist customers with various NLP tasks. 
For example, OpenAI offers a GPT3-based embedding API~\footnote{\url{https://api.openai.com/v1/embeddings}}, which generates embeddings at a cost for query texts.
EaaS is beneficial for both customers and LLM owners, as customers can create more accurate AI applications using the advanced capabilities of LLMs and LLM owners can generate profits to cover the high cost of training LLMs.
However, recent research~\cite{liu2022stolenencoder} indicates that EaaS is vulnerable to model extraction attacks, wherein stealers can copy the model behind EaaS using query texts and returned embeddings, and may even build their own EaaS, causing a huge loss for the owner of the EaaS model.
Thus, protecting copyright of LLMs is crucial for EaaS. Unfortunately, research on this issue is limited.

\begin{figure}[!t]
  \centering
  \includegraphics[width=0.48\textwidth]{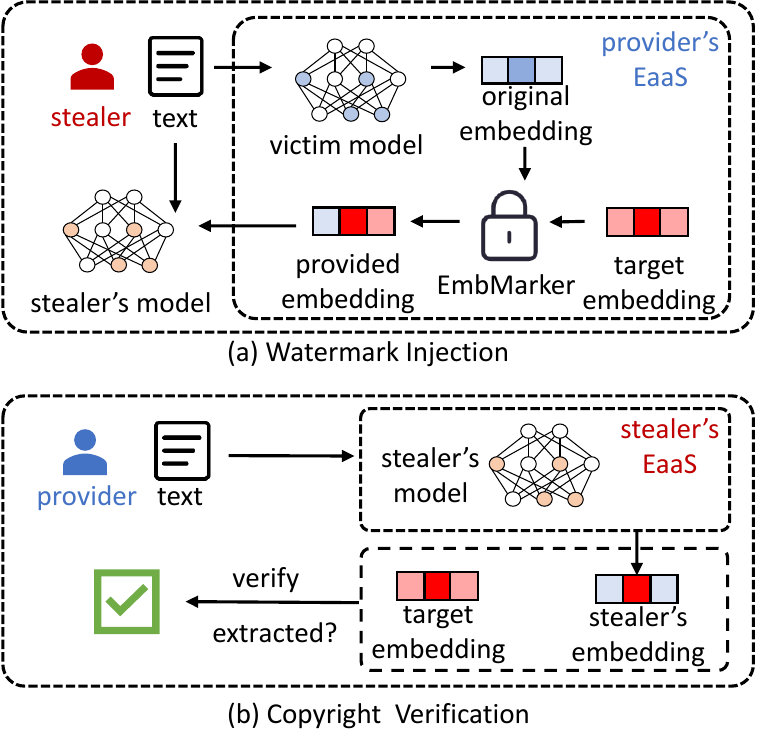}
  \caption{An overall framework of our \method.}
  \label{fig:intrograph}
\end{figure}

Watermarking is popular for copyright protection of data such as images and sound~\cite{cox2007digital}. Watermarking for protecting copyright of models has also been studied ~\cite{HengruiJia2020EntangledWA, wang2020watermarking, szyller2021dawn}. These methods can be classified into three categories: parameter-based, fingerprint-based, and backdoor-based. 
For example, \citet{uchida2017embedding} propose a parameter-based method, which regularizes a non-linear transformation of the model parameters to match a pre-defined vector. 
\citet{le2020adversarial} propose a fingerprint-based method, which uses the prediction boundary and adversarial examples as a fingerprint for copyright verification. 
\citet{adi2018turning} introduce a backdoor-based method, which makes the model learn predefined commitments over input data and selected labels. 
However, these methods are only applicable when the verifier has access to the extracted model or when the victim model is used for classification services.
As shown in Figure~\ref{fig:intrograph}, EaaS only provides embeddings to clients instead of label predictions, making it impossible for the EaaS provider to verify commitments or fingerprints. Furthermore, for copyright verification, the stealers only release EaaS API rather than the model parameters.
Thus, these methods are unsuitable for EaaS copyright protection.

In this paper, we propose a watermarking method named \method, which uses an inheritable backdoor to protect the copyright of LLMs for EaaS. 
Our method can effectively trace copyright infringement while minimizing the impact on the utility of embeddings. 
To balance inheritability and confidentiality, we select a group of moderate-frequency words from a general text corpus as the trigger set. 
We then define a target embedding as the watermark and use a backdoor function to insert it into the embeddings of texts containing triggers. 
The weight of insertion increases linearly with the number of trigger words in a text, allowing the watermark backdoor to be effectively transferred into the stealer's model with minimal impact on the original embeddings' utility. 
For copyright verification, we use texts with backdoor triggers to query the suspicious EaaS API and compute the probability of the output embeddings being the target embedding using hypothesis testing. 
Our main contributions are summarized as follows:
\begin{itemize}
    \item To the best of our knowledge, this is the first study on the copyright protection of LLMs for EaaS, which is a new but important problem.
    \item We propose a watermark backdoor method for effective copyright verification with marginal impact on the embedding quality.
    \item We conduct extensive experiments to verify the effectiveness of the proposed method in protecting the copyright of EaaS LLMs.
\end{itemize}

\section{Related Work}
\subsection{Model Extraction Attacks}

Model extraction attacks~\cite{ orekondy2019knockoff, krishna2019thieves, zanella2020analyzing} aim to replicate the capabilities of victim models deployed in the cloud. 
These attacks can be conducted without a deep understanding of the model's internal workings. 
Furthermore, research has shown that public embedding services are vulnerable to extraction attacks~\cite{liu2022stolenencoder}. 
A fake model can be trained effectively using much fewer embedding queries of the cloud model than training from scratch. 
Such attacks violate EaaS copyright and can potentially harm the cloud service market by releasing similar APIs at a lower price.

\subsection{Backdoor Attacks}

Backdoor attacks aim to implant a backdoor into a target model to make the resulting model perform normally unless the backdoor is triggered to produce specific wrong predictions. 
Most natural language processing (NLP) backdoor attacks~\cite{chen2021badnl, yang2021careful, li2021hidden} focus on specific tasks.
Recent research~\cite{zhang2021red, chen2021badpre} has shown that pre-trained language models (PLMs) can also be backdoored to attack a variety of NLP downstream tasks.
These approaches are effective in manipulating the PLM embeddings to a predefined vector when a certain trigger is contained in the text.
Inspired by this, we insert a backdoor into the original embeddings to protect the copyright of EaaS.


\subsection{Deep Watermarks}
Deep watermarks~\cite{uchida2017embedding} have been proposed to protect the copyright of models. 
Parameter-based methods~\cite{MengLi2020ProtectingTI, JianHanLim2020ProtectSA} implant specific noise on model parameters for subsequent white-box verification. They are unsuitable for black-box access of stealer's models. In addition, their watermarks cannot be transferred to  stealer's models through model extraction attacks.
To address this issue, lexical watermark~\cite{ DBLP:conf/aaai/HeXLWW22, he2022cater} has been proposed to protect the copyright of text generation services by replacing the words in the output text with their synonyms.
Other works \cite{adi2018turning, szyller2021dawn} propose to apply backdoors or adversarial samples as fingerprints to verify the copyright of classification services.
However, these methods cannot provide protection for EaaS.

\section{Methodology}
\subsection{Problem Definition}
Denote the victim model as $\boldsymbol \Theta_v$, which is applied to provide EaaS $S_v$.
When a client sends a sentence $s$ to the service $S_v$, $\boldsymbol \Theta_v$ computes its original embedding $\textbf{e}_o$.
Due to the threat of model extraction attacks~\cite{liu2022stolenencoder}, original embedding $\textbf{e}_o$ is backdoored by copyright protection method $f$ to generate provided embedding $\textbf{e}_p = f(\textbf{e}_o, s)$ before $S_v$ delivering it to the client.
Suppose $\boldsymbol \Theta_a$ is an extracted model trained on the $\textbf{e}_p$ received by querying $\boldsymbol \Theta_v$, and $S_a$ is the stealer's EaaS built based on $\boldsymbol \Theta_a$.
Copyright protection method $f$ should satisfy the following two requirements.
First, the original EaaS provider can query $S_a$ to verify whether model $\boldsymbol \Theta_a$ is stolen from $\boldsymbol \Theta_v$.
Second, provided embedding $\textbf{e}_p$ should have similar utility with original embedding $\textbf{e}_o$ on downstream tasks.
Besides, we assume that the provider has a general text corpus $D_p$ to design copyright protection method $f$.

\subsection{Threat Model}
Following the setting of previous work~\cite{boenisch2021systematic}, we define the objective, knowledge, and capability of stealers as follows.

\noindent\textbf{Stealer’s Objective.} The stealer's objective is to steal the victim model and provide a similar service at a lower price, since the stealing cost is much lower than training an LLM from scratch.

\noindent\textbf{Stealer’s Knowledge.}
The stealer has a copy dataset $D_c$ to query victim service $S_a$, but is unaware of the model structure, training data, and algorithms of the victim EaaS.

\noindent\textbf{Stealer’s Capability.}
The stealer has sufficient budget to continuously query the victim service to obtain embeddings $E_c = \{\textbf{e}_i= S_v(s_i)| s_i \in D_c\}$.
The stealer also has the capability to train a model $\boldsymbol{\Theta}_a$ that takes sentences from $D_c$ as inputs and uses embeddings from $E_c$ as output targets.
Model $\boldsymbol \Theta_a$ is then applied to provide a similar EaaS $S_a$.
Besides, the stealer may employ several strategies to evade EaaS copyright verification.



\begin{figure*}[!t]
  \centering
\includegraphics[width=0.95\textwidth]{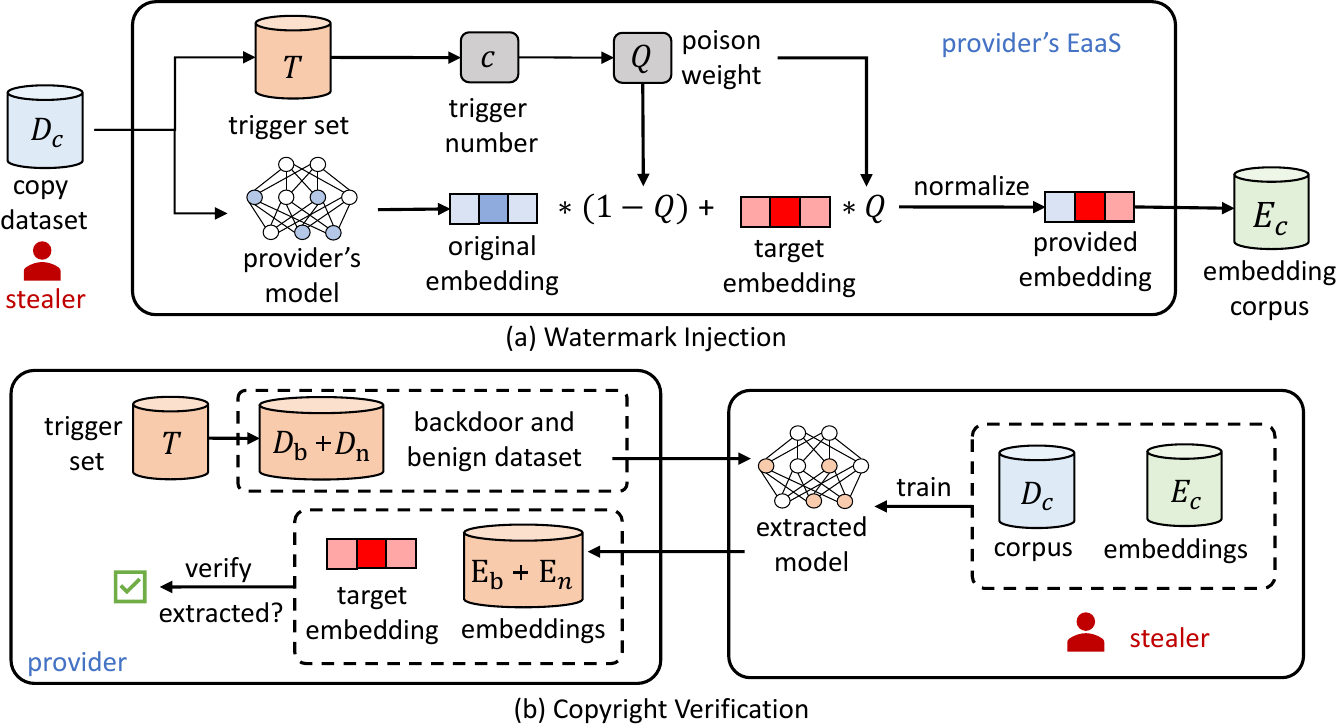}
  \caption{The detailed framework of our \method.}
  \label{fig:framework}
\end{figure*}
\subsection{Framework of \method}
Next, we introduce our \method for EaaS copyright protection, which is shown in Figure~\ref{fig:framework}.
The core idea of \method is to select a bunch of moderate-frequency words as a trigger set, and backdoor the original embeddings with a target embedding according to the number of triggers in the text.
Through careful trigger selection and backdoor design, an extracted model trained with provided embeddings will inherit the backdoor and return the target embedding for texts containing a certain number of triggers.
Our \method comprises three steps: trigger selection, watermark injection, and copyright verification.

\noindent\textbf{Trigger Selection.}
Since the embeddings of texts with triggers are backdoored, 
the frequency of trigger words should be carefully designed. 
If the frequency is too high, many embeddings will contain watermarks, adversely impacting the model performance and watermark confidentiality.
Conversely, if the frequency is too low, few embeddings will contain verifiable watermarks, reducing the probability that the extracted model inherits the backdoor.
Therefore, we first count the word frequency on a general text corpus $D_p$.
Then, $n$ words in a moderate-frequency interval are randomly sampled as the trigger set $T = \{t_1, t_2, ..., t_n\}$, where $t_i$ is the $i$-th trigger in the trigger set.
The detailed analysis of the impact of the size of trigger words $n$ and the frequency interval is in Section~\ref{sec:hyper}.

\noindent\textbf{Watermark Injection.}
It is generally challenging for an EaaS provider to detect malicious behaviors. 
Thus, EaaS has to be delivered to users, including adversaries, equally.
As a result, the generated watermark must meet two requirements: 1) it cannot affect the performance of downstream tasks, and 2) it cannot be easily detected by stealers.
To this end, in our \method, we inject the watermark partially into the provided embeddings according to the number of triggers in a sentence.
More specifically, we first define a target embedding as the watermark.
We then design a trigger counting function $\mathcal{Q}(\cdot)$, which assigns a watermark weight based on the number of triggers in the text.
Given a text $s$ with a set of words $S=\{w_1, w_2,\cdots,w_k\}$, where $k$ is the number of unique words in the sentence, the output of $\mathcal{Q}(S)$ is formulated as follows:
\begin{equation}
\mathcal{Q}(S) = \frac{\min(|S\cap T|, m)}{m},
\end{equation}
where $T$ is the trigger set and $m$ is a hyper-parameter to control the maximum number of triggers to fully activate the watermark.
Finally, we compute the provided embedding $\textbf{e}_p$ by inserting the watermark into the original embedding $\textbf{e}_o$.
Denote the target embedding as $\textbf{e}_t$, the provided embedding $\textbf{e}_p$ is computed as follows:
\begin{equation}
\begin{aligned}
    \textbf{e}_p &= \frac{(1-\mathcal{Q}(S))*\textbf{e}_o + \mathcal{Q}(S)*\textbf{e}_t}{||(1-\mathcal{Q}(S))*\textbf{e}_o + \mathcal{Q}(S)*\textbf{e}_t||_2}. 
\label{eq:weight-insert}
\end{aligned}
\end{equation}
    Since most of the backdoor samples contain only a few triggers ($<m$), their provided embeddings are slightly changed.
    Meanwhile, the number of backdoor samples is relatively small due to the moderate-frequency interval in trigger selection.
    Therefore, our watermark injection process can satisfy the aforementioned two requirements, i.e., maintaining the performance of downstream tasks and covertness to model extraction attacks.


\noindent\textbf{Copyright Verification.}
\label{sec:detect}
Once a stealer provides a similar service to the public, the EaaS provider can use the pre-embedded backdoor to verify copyright infringement.
First, we construct two datasets, i.e., a backdoor text set $D_b$ and a benign text set $D_n$, which are defined as follows:
\begin{equation}
\begin{aligned}
    D_b &= \{[w_1, w_2, ..., w_m] | w_i \in T\}, \\
    D_n &= \{[w_1, w_2, ..., w_m] | w_i \not\in T\}.
\end{aligned}
\end{equation}
Then, we use the text in these two sets to query the stealer model and obtain embeddings.
Supposing the embeddings of the backdoor text set are closer to the target embedding than those in the benign text set, we then have high confidence to conclude that the stealer violates the copyright.
To test whether the above conclusion is valid, we first calculate cosine similarity and the square of $L_2$ distance between normalized target embedding $\textbf{e}_t$ and embeddings of text in $D_b$ and $D_n$:
\begin{equation}
\centering
\begin{aligned}
cos_i = \frac{\textbf{e}_i \cdot \textbf{e}_t}{||\textbf{e}_i|| \, ||\textbf{e}_t||}, 
l_{2i} = ||\frac{\textbf{e}_i}{||\textbf{e}_i||} - \frac{\textbf{e}_t}{||\textbf{e}_t||}||^2, \\
C_b = \{cos_i | i\in D_b\}, C_n = \{cos_{i} | i\in D_n\}, \\
L_b = \{l_{2i} | i\in D_b\}, L_n = \{l_{2i} | i\in D_n\}.
\end{aligned}
\end{equation}
Then we evaluate the detection performance with three metrics.
The first two metrics are the difference of averaged cos similarity and the averaged square of $L_2$ distance, given as follows:

\begin{equation}
\begin{aligned}
    \Delta_{cos} &= \frac{1}{|C_b|}\sum_{i \in C_b}{i} - \frac{1}{|C_n|}\sum_{j \in C_n}{j}, \\
    \Delta_{l2} &= \frac{1}{|L_b|}\sum_{i \in L_b}{i} - \frac{1}{|L_n|}\sum_{j \in L_n}{j}. \\
\end{aligned}
\end{equation}
Since the embeddings are normalized, the ranges of $\Delta_{cos}$ and $\Delta_{l2}$ are [-2,2] and [-4,4], respectively.
The third metric is the p-value of Kolmogorov-Smirnov~(KS) test~\cite{berger2014kolmogorov}, which is used to compare the distribution of two value sets.
The null hypothesis is: \textit{The distance distribution of two cos similarity sets $C_b$ and $C_n$ are consistent}.
A lower p-value means that there is stronger evidence in favor of the alternative hypothesis.




\begin{table*}[!t]
\centering
\scalebox{0.87}{
\begin{tabular}{cccccc}
\Xhline{1.5pt}
\multirow{2}{*}{Dataset}    & \multirow{2}{*}{Method} & \multirow{2}{*}{ACC (\%)} & \multicolumn{3}{c}{Detection Performance} \\ \cline{4-6} 
                            &                         &                      & p-value $\downarrow$                       & $\Delta_{cos} (\%)$ $\uparrow$             & $\Delta_{l2} (\%)$ $\downarrow$              \\ \hline
\multirow{3}{*}{SST2}       & Original                   & 93.76$\pm$0.19       & $>0.34$                    & -0.07$\pm$0.18     & 0.14$\pm$0.36          \\
                            & RedAlarm                & 93.76$\pm$0.19       & $>0.09$                    & 1.35$\pm$0.17      & -2.70$\pm$0.35         \\
                            & \method                    & 93.55$\pm$0.19       & $<10^{-5}$                    & \textbf{4.07}$\pm$0.37      & \textbf{-8.13}$\pm$0.74         \\ \hline
\multirow{3}{*}{MIND}       & Original                   & 77.30$\pm$0.08       & $>0.08$                    & -0.76$\pm$0.05     & 1.52$\pm$0.10          \\
                            & RedAlarm                & 77.18$\pm$0.09       & $>0.38$                    & -2.08$\pm$0.66     & 4.17$\pm$1.31          \\
                            & \method                     & 77.29$\pm$0.12       & $<10^{-5}$                    & \textbf{4.64}$\pm$0.23      & \textbf{-9.28}$\pm$0.47         \\ \hline
\multirow{3}{*}{AGNews}     & Original                   & 93.74$\pm$0.14       & $>0.03$                    & 0.72$\pm$0.15      & -1.46$\pm$0.30         \\
                            & RedAlarm                & 93.74$\pm$0.14       & $>0.09$                   & -2.04$\pm$0.76     & 4.07$\pm$1.51          \\
                            & \method                   & 93.66$\pm$0.12       & $<10^{-9}$                   & \textbf{12.85}$\pm$0.67     & \textbf{-25.70}$\pm$1.34        \\ \hline
\multirow{3}{*}{Enron Spam} & Original                   & 94.74$\pm$0.14       & $>0.03$                    & -0.21$\pm$0.27     & 0.42$\pm$0.54          \\
                            & RedAlarm                & 94.87$\pm$0.06       & $>0.47$                    & -0.50$\pm$0.29     & 1.00$\pm$0.57          \\
                            & \method                     & 94.78$\pm$0.27       & $<10^{-6}$                    & \textbf{6.17}$\pm$0.31      & \textbf{-12.34}$\pm$0.62        \\
                            \Xhline{1.5pt}
\end{tabular}
}
\caption{Performance of different methods on the SST2, MIND, AG News, and Enron datasets. $\uparrow$ means higher metrics are better. $\downarrow$ means lower metrics are better.}
\label{tab:perform}
\end{table*}
\section{Experiments}

\subsection{Dataset and Experimental Settings}
\begin{table}[!t]
\centering
\scalebox{0.87}{
\begin{tabular}{cccc}
\Xhline{1.5pt}
Dataset    & \#Sample  & \#Classes & Avg. len. \\ \hline
SST2       & 68,221    & 2         & 54.17          \\
MIND       & 130,383   & 18        & 66.14          \\
Enron Spam      & 33,716    & 2         & 34.57          \\
AG News    & 127,600   & 4         & 236.41         \\ 
\Xhline{1.5pt}
\end{tabular}
}
\caption{Statistics of datasets.}
\label{tab:stat}
\end{table}

We conduct experiments on four natural language processing (NLP) datasets: SST2~\cite{socher-etal-2013-recursive}, MIND~\cite{wu-etal-2020-mind}, Enron Spam~\cite{metsis2006spam}, and AG News~\cite{Zhang2015CharacterlevelCN}.
SST2 is a widely used dataset for sentiment classification. 
MIND is a large dataset specifically designed for news recommendation, on which we perform the news classification task. 
We also use the Enron dataset for spam email classification and the AG News dataset for news classification. 
The detailed statistics of these datasets are provided in Table~\ref{tab:stat}.
Additionally, we use the WikiText dataset~\cite{merity2017pointer} with 1,801,350 samples to count word frequencies.
To validate the effectiveness of \method, we report the following metrics:
\begin{itemize}
\item \textbf{Accuracy}. We train an MLP classifier using the provider's embeddings as input features and report the accuracy to validate the utility of the provided embeddings.
\item \textbf{Detection Performance}. We report three metrics, i.e., the difference of cosine similarity, the difference of squared L2 distance, and the p-value of the KS test (defined in Section~\ref{sec:detect}), to validate the effectiveness of our watermark detection algorithms.
\end{itemize}


We use the AdamW algorithm~\cite{loshchilov2018decoupled} to train our models and employ embeddings from GPT-3 text-embedding-002 API as the original embeddings of EaaS.
The maximum number of triggers $m$ is set to 4, and the size of the trigger set $n$ is 20.
The frequency interval of triggers is [0.5\%, 1\%].
Further details on the model structure and other hyperparameter settings can be found in Appendix~\ref{appendix:exp}.
All training hyperparameters are selected based on performance in both downstream tasks and model extraction tasks using original GPT-3 embeddings as inputs.
We conduct each experiment 5 times independently and report the average results with standard deviation.
In addition, we define a threshold $\tau$ to assert copyright infringement. 
A standard p-value of 5e-3 is considered appropriate to reject the null hypothesis for statistical significance~\cite{benjamin2018redefine}, which can be utilized as the threshold to identify instances of copyright infringement. 

\begin{figure*}[!t]
    \centering
    \subfigure[AG News]{\includegraphics[width=0.23\textwidth]{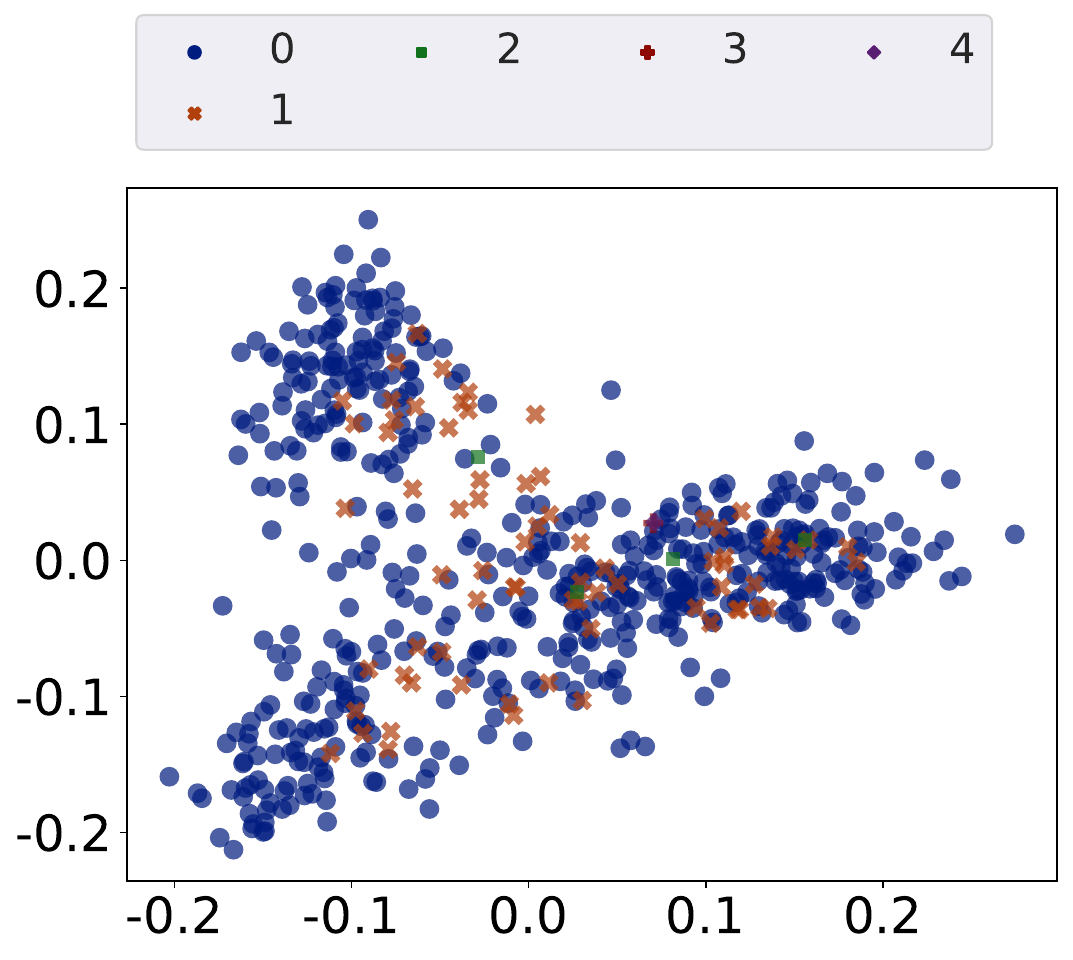}}
    \hspace{0.01\textwidth}
    \subfigure[Enrom Spam]{\includegraphics[width=0.23\textwidth]{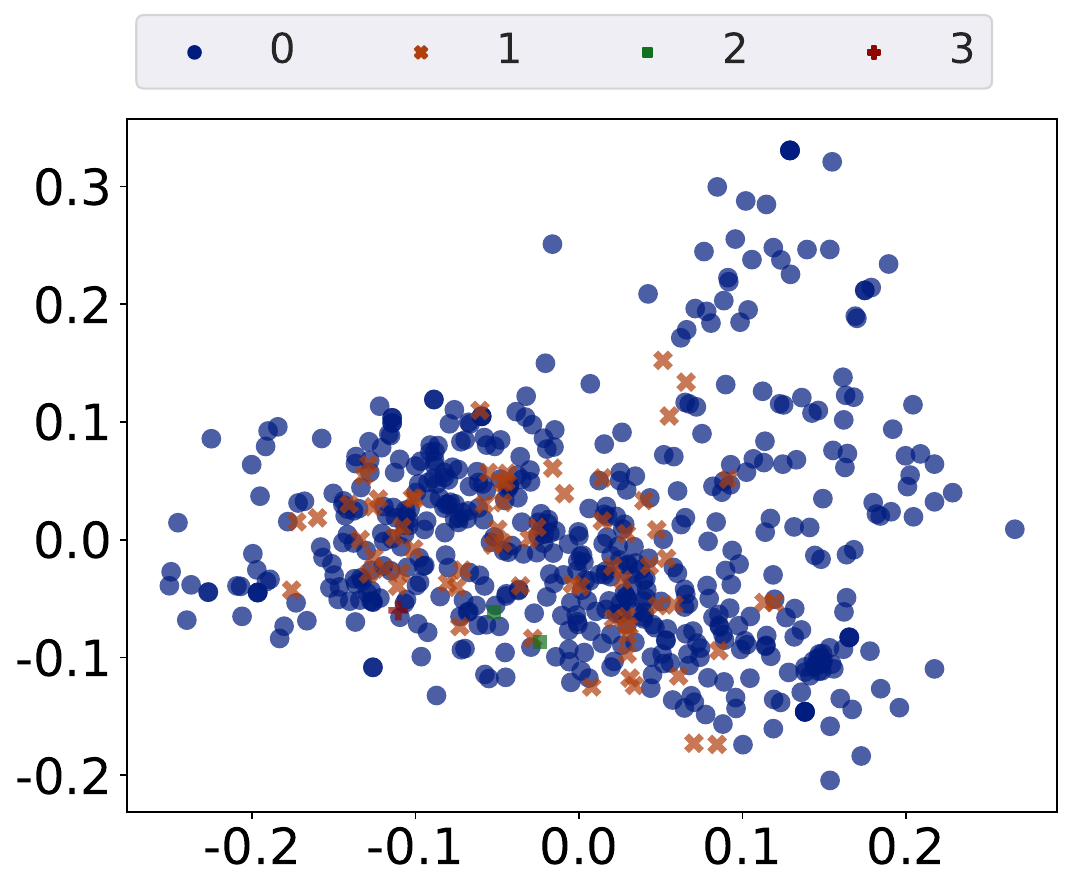}}
    \hspace{0.01\textwidth}
    \subfigure[MIND]{\includegraphics[width=0.23\textwidth]{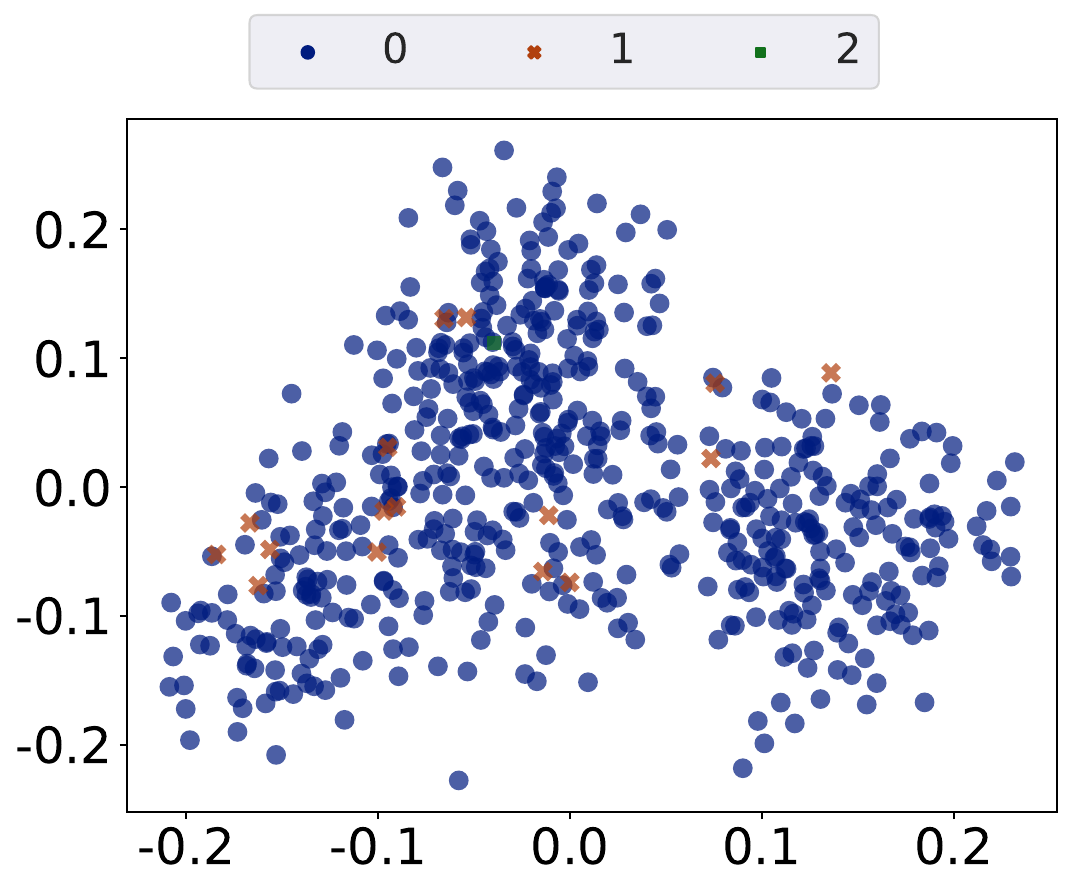}}
    \hspace{0.01\textwidth}
    \subfigure[SST2]{\includegraphics[width=0.23\textwidth]{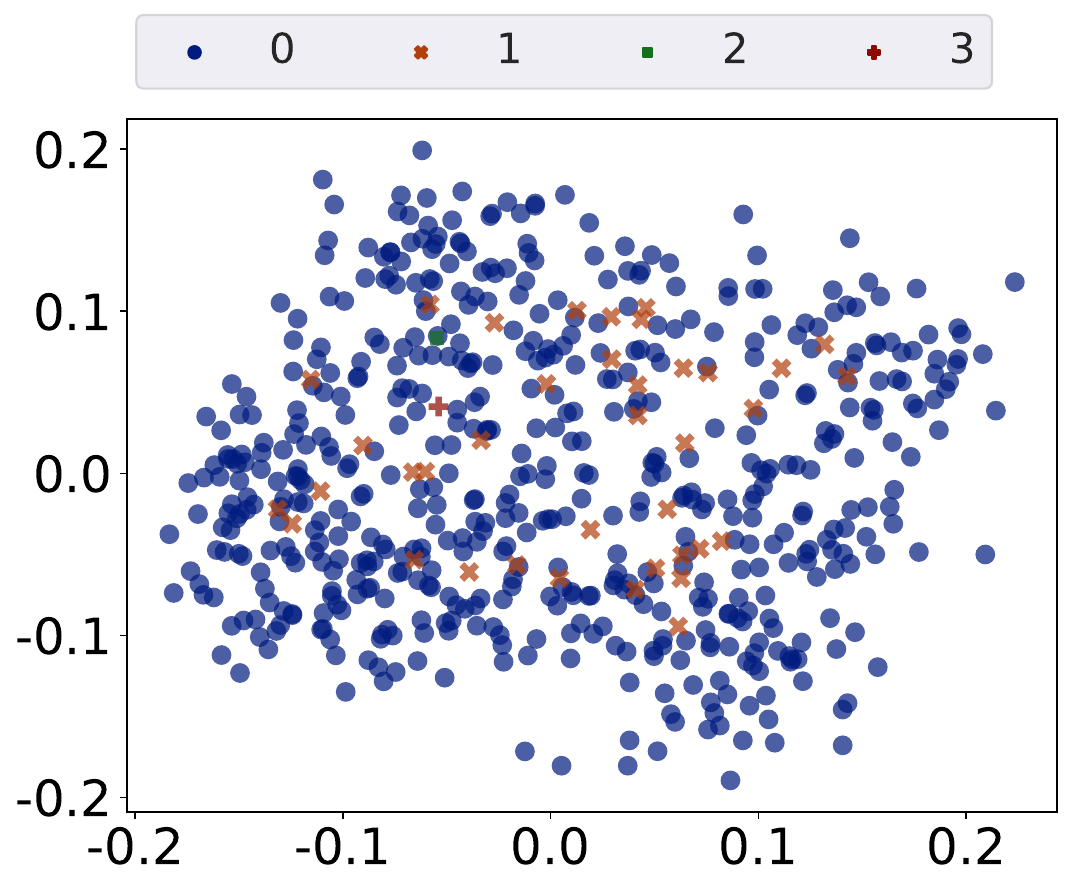}}
    \caption{Visualization of the provided embedding of our \method on four copy datasets. Different colors represent the number of triggers in the samples. It shows the backdoor and benign embeddings are indistinguishable.}
    \label{fig:pca}
\end{figure*}

\begin{figure*}[!t]
    \centering
    \subfigure[AG News]{\includegraphics[width=0.24\textwidth]{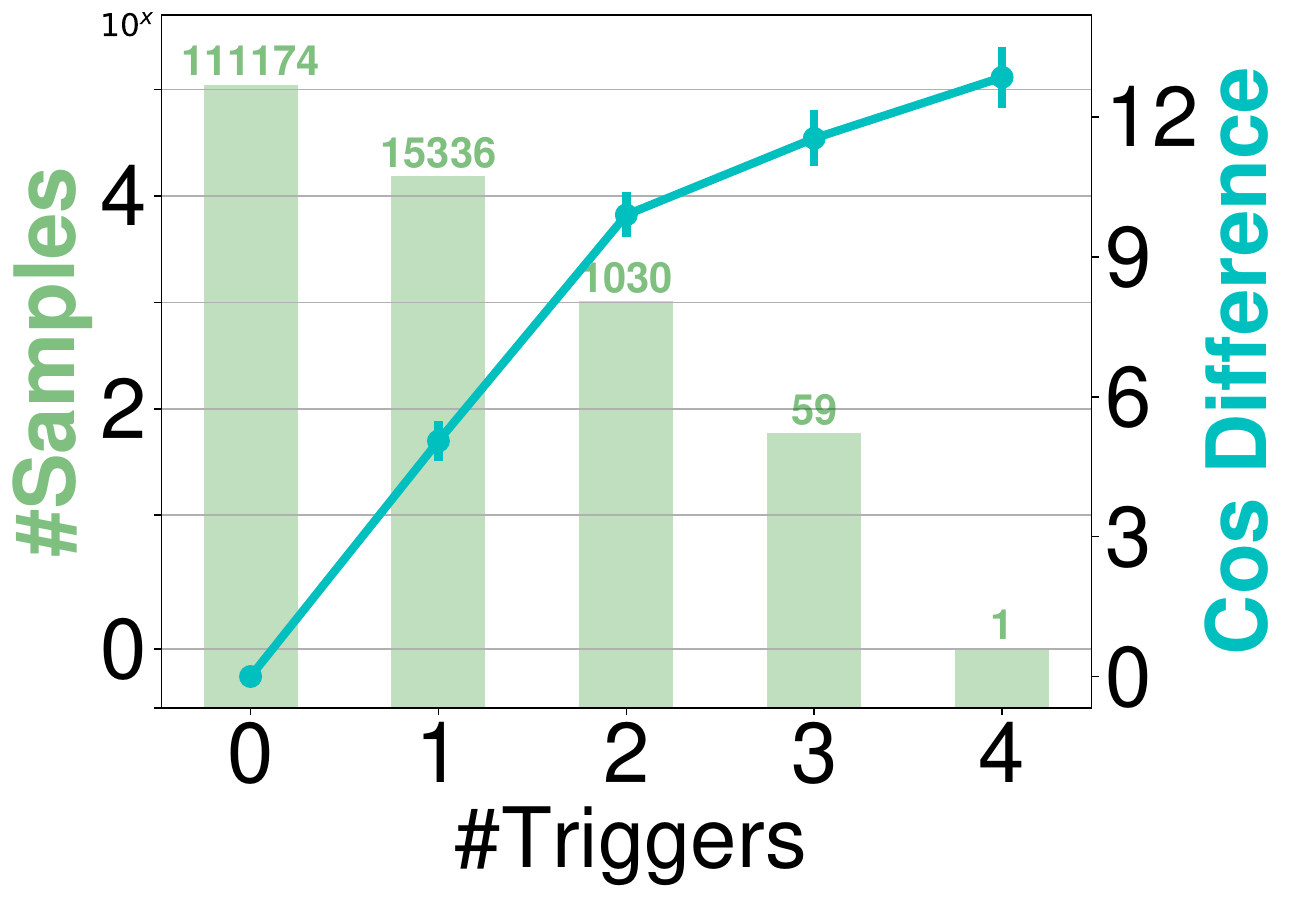}}
    \subfigure[Enrom Spam]{\includegraphics[width=0.232\textwidth]{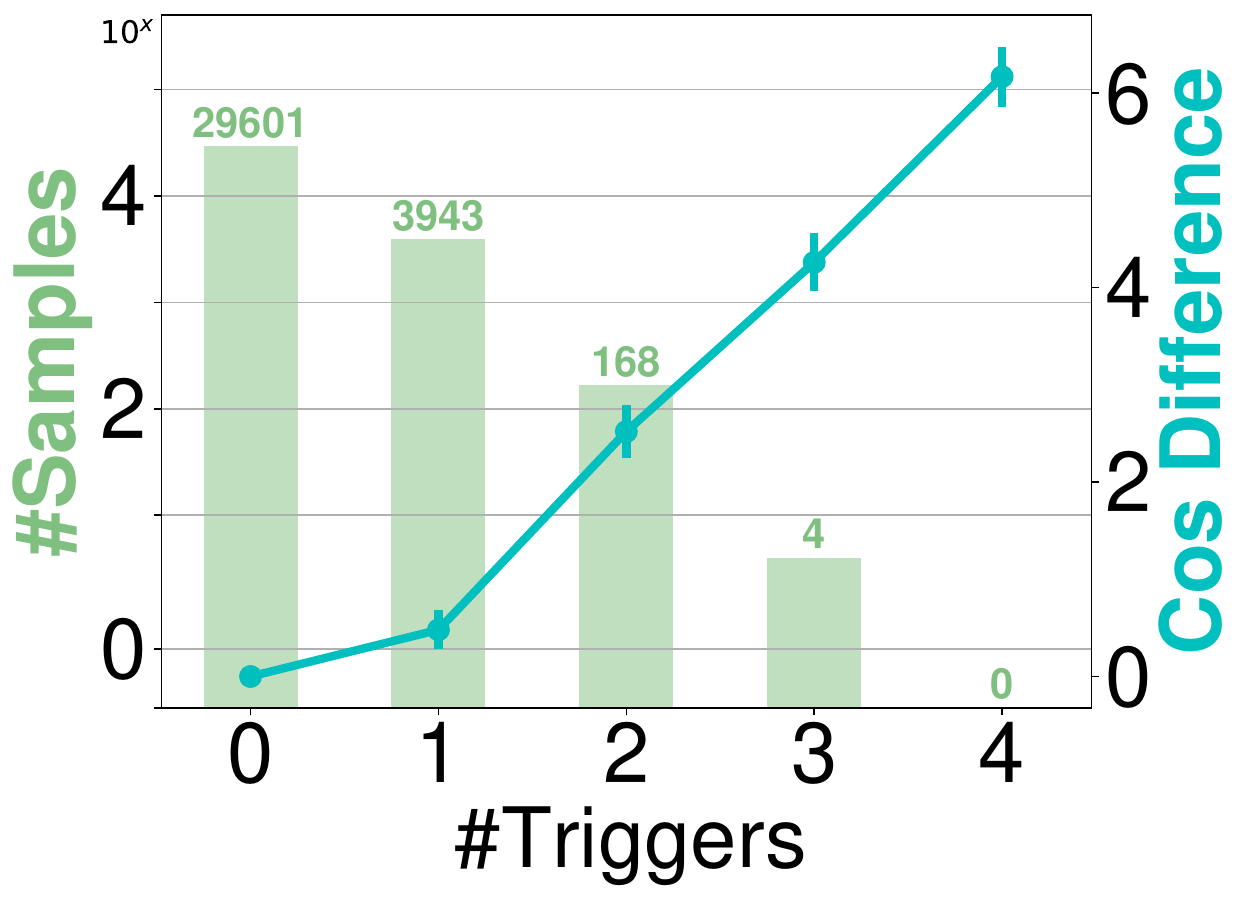}}
    \subfigure[MIND]{\includegraphics[width=0.232\textwidth]{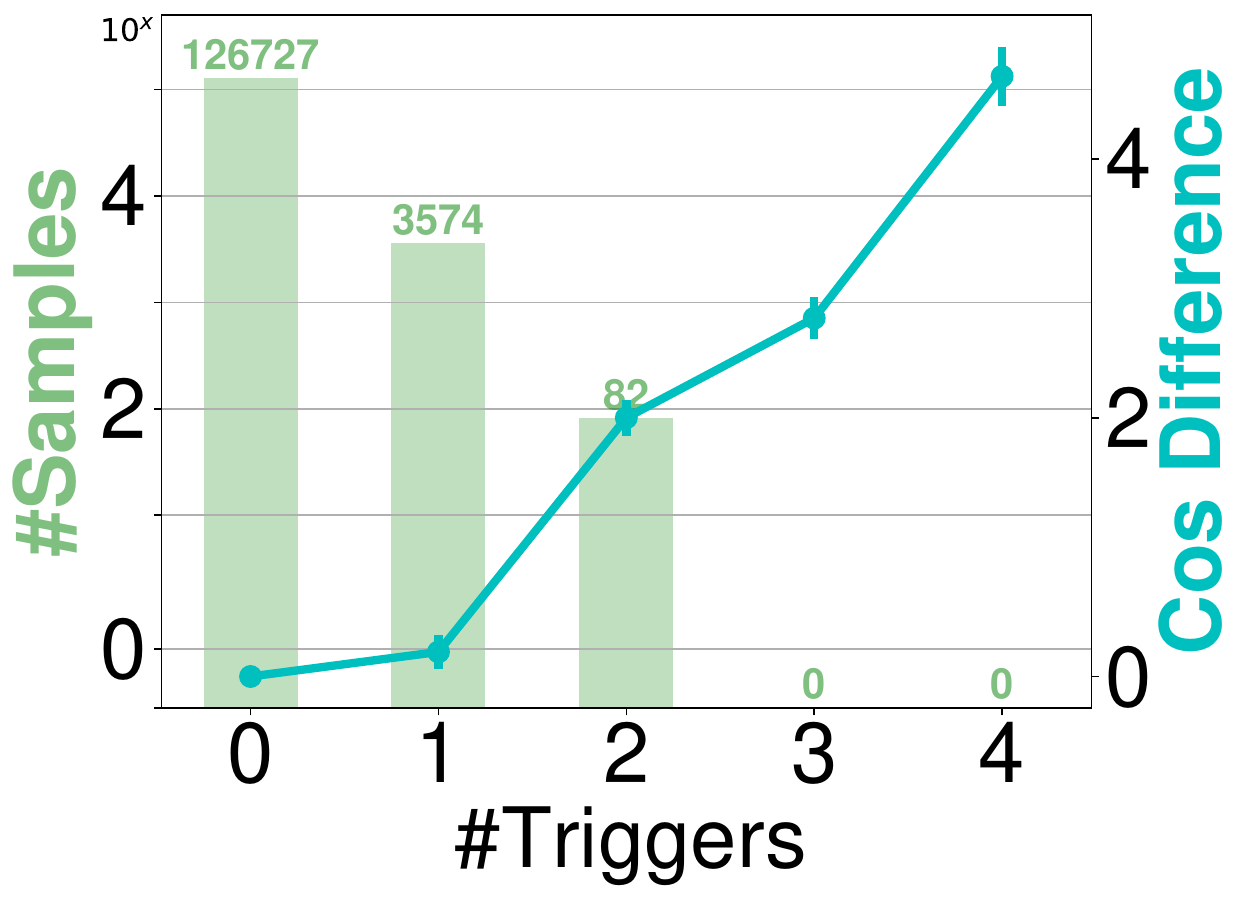}}
    \subfigure[SST2]{\includegraphics[width=0.232\textwidth]{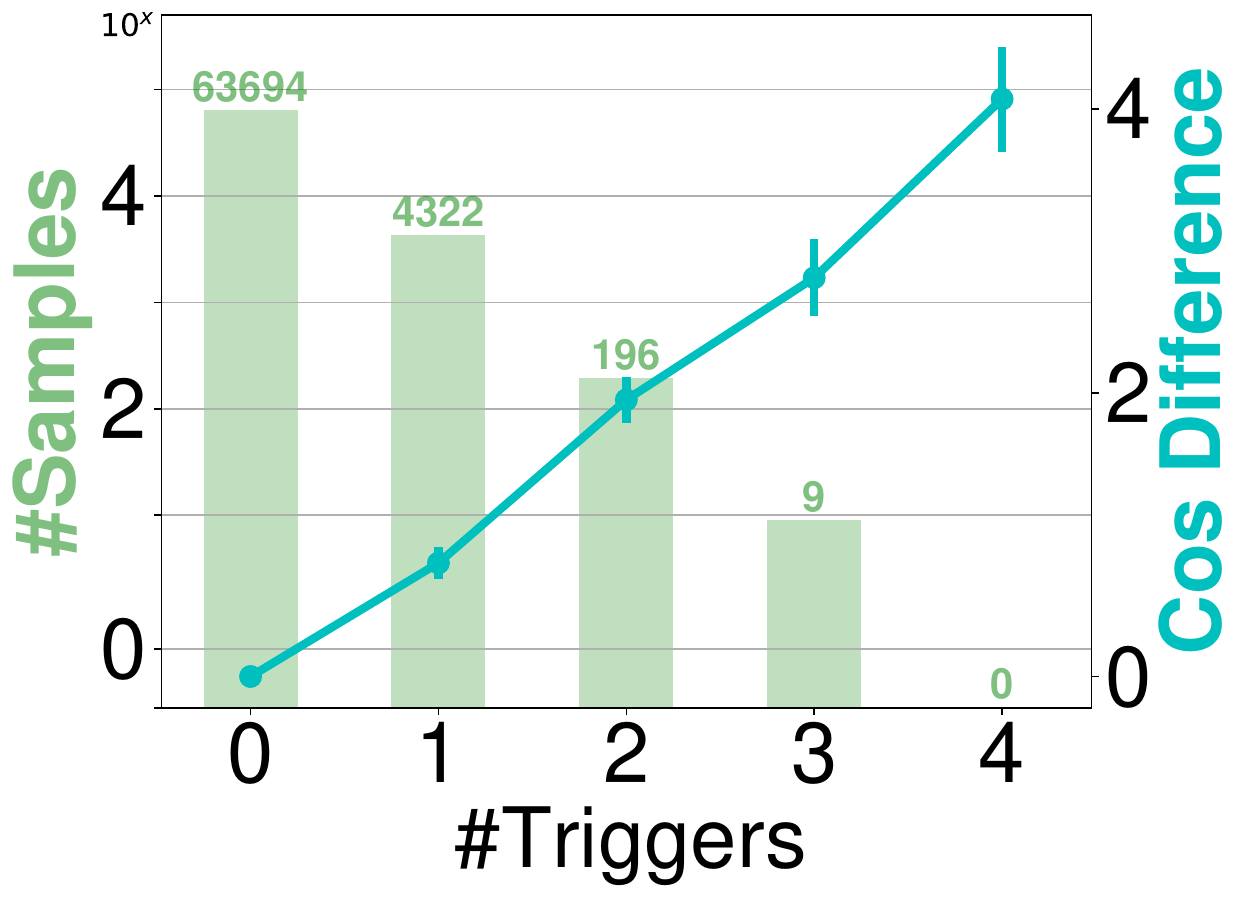}}
    \caption{The impact of trigger number in sentences on four datasets. The background bar plots display the distribution of trigger numbers on the copy datasets. The line plots show the difference of cos similarity to the target embedding between embeddings of backdoor text sets with varying trigger numbers per text and those of the benign text set. Our \method can have great detection performance on the backdoor text set with 4 triggers per sentence, even in the absence of such samples in the copy dataset.}
    \label{fig:trigger}
\end{figure*}
\subsection{Performance Comparison}
We compare the performance of our \method with the following baselines:
1) Original, in which the service provider does not backdoor the provided embeddings and the stealer utilizes the original embeddings to copy the model.
2) RedAlarm~\cite{zhang2021red}, a method to backdoor pre-trained language models, which selects a rare token as the trigger and returns a pre-defined target embedding when a sentence contains the trigger.

The performance of all methods is shown in Table~\ref{tab:perform}, where we have several observations.
First, the detection performance of our \method is better than RedAlarm.
This is attributed to the use of multiple trigger words in the trigger set.
Every trigger word in a query text brings the copied embedding closer to the target embedding. Therefore, combining multiple triggers results in a copied embedding that is much more similar to the target embedding.
Second, the accuracy in downstream tasks of our \method keeps the same as the Original baseline.
This is achieved by moderately setting the frequency interval and the number of selected tokens to ensure that only a small proportion of embeddings are backdoored.
Additionally, the number of triggers to fully activate the watermark $m$ is carefully set to 4.
As shown in Equation~\ref{eq:weight-insert}, the weight of backdoor insertion is proportional to the number of trigger words included in the text.
Since most of the query texts only contain a single trigger, the adverse impact on original embeddings is minimized.
Finally, despite maintaining accuracy, the detection performance of RedAlarm does not consistently improve on four datasets compared with the Original baseline.
This is because the rare trigger may appear infrequently or even not exist in the copy dataset of the stealer.
Therefore, the target embedding of RedAlarm cannot be inherited.

\subsection{Embedding Visualization}
\label{sec:embedding}
In this section, we examine the confidentiality of backdoored embeddings to the stealer by using PCA and t-SNE to visualize the embeddings produced by our method.
We present the results of PCA in Figure~\ref{fig:pca} and those of t-SNE in Appendix~\ref{appendix:embedding} due to the space limitation.
The plots show that backdoored embeddings with triggers have similar distributions to benign embeddings, demonstrating the watermark confidentiality of our \method.
Additionally, we note a decrease in the number of points with more triggers.
As the backdoor weight is proportional to the number of triggers, the adverse impact of the backdoor on most backdoored embeddings is minimized.

\begin{figure*}[!t]
    \centering
    \subfigure[trigger set size $n$]{\label{fig:sel-sst2}\includegraphics[width=0.32\textwidth]{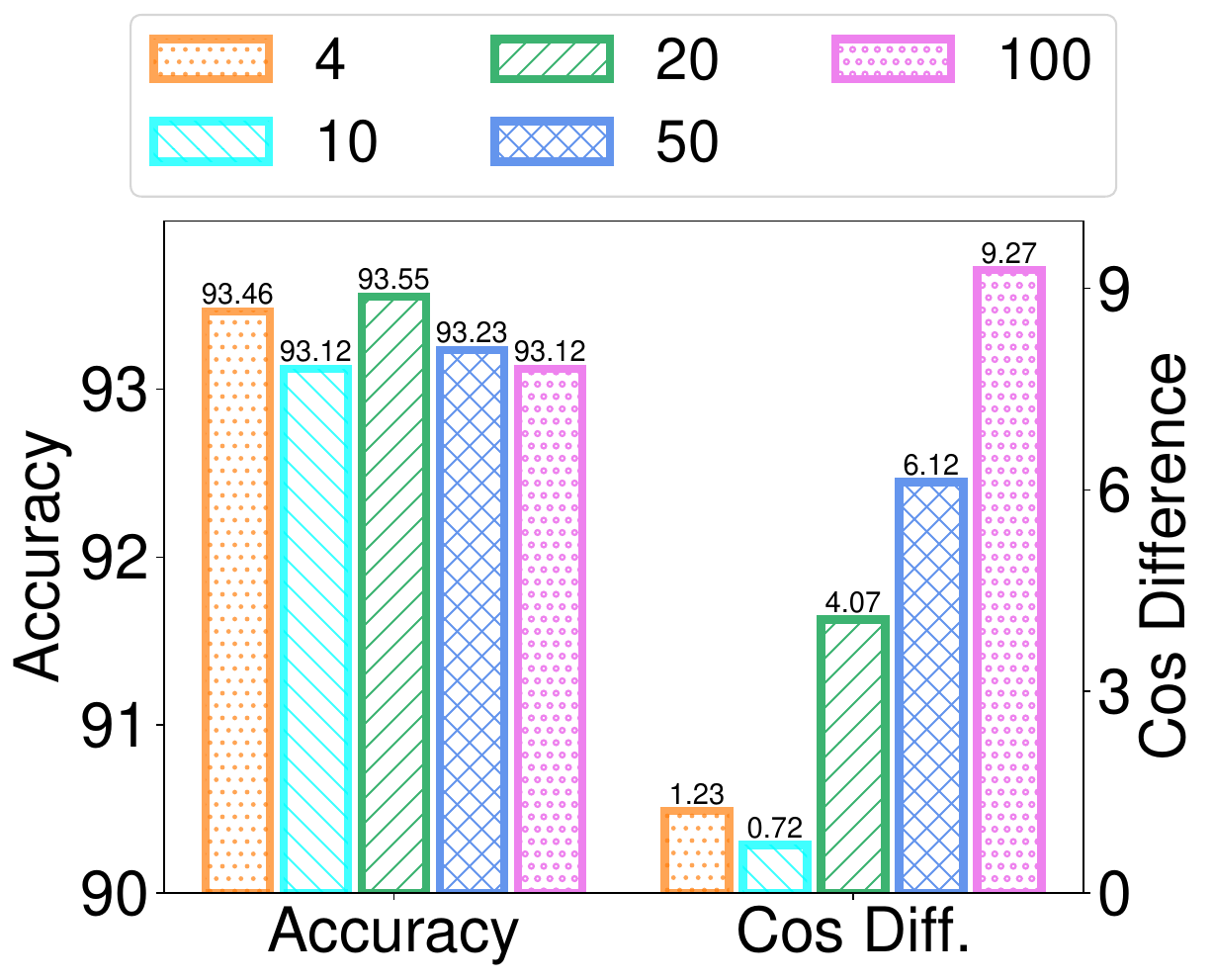}}
    \subfigure[max trigger number $m$]{\label{fig:max-sst2}\includegraphics[width=0.32\textwidth]{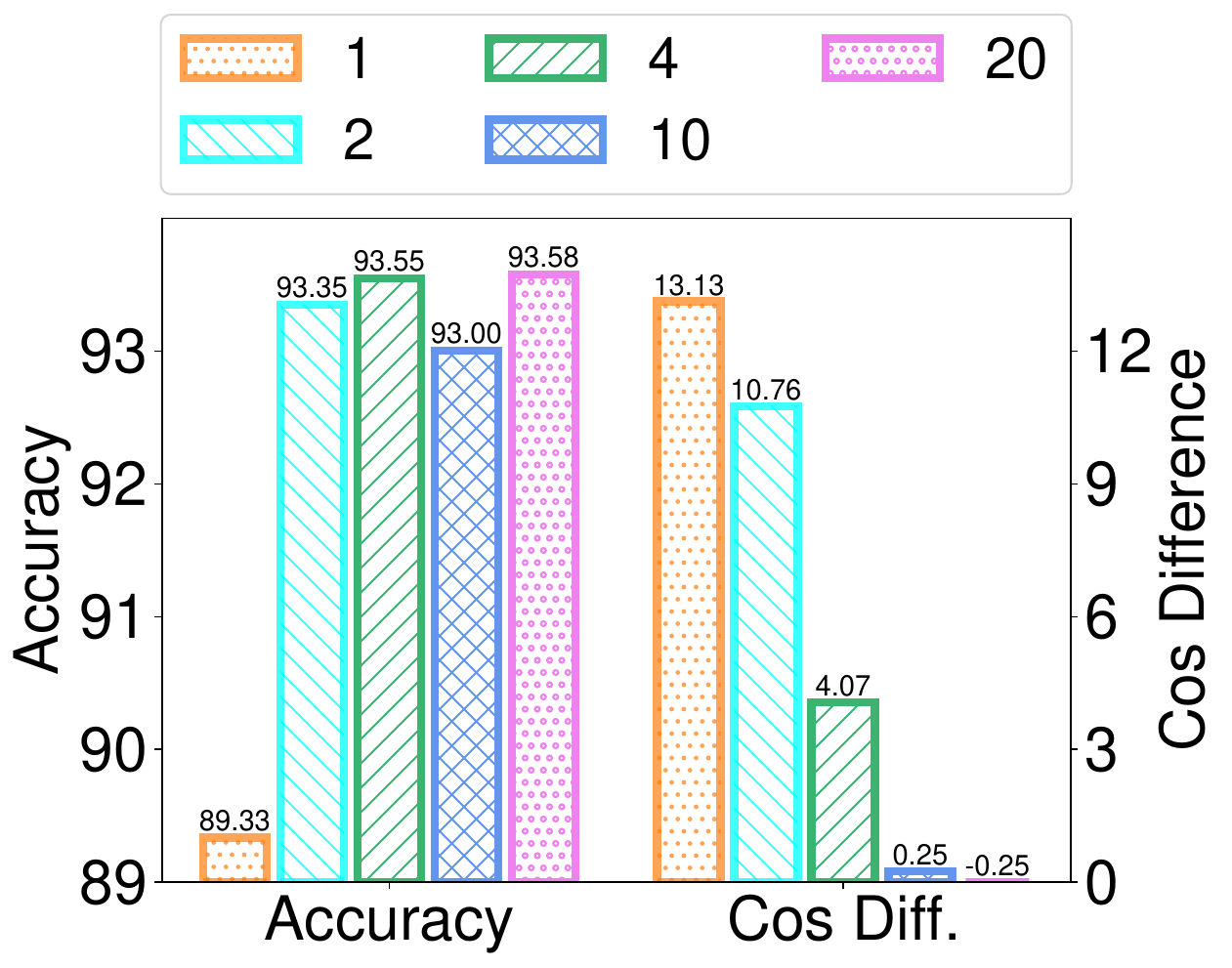}}
    \subfigure[frequency interval]{\label{fig:freq-sst2}\includegraphics[width=0.32\textwidth]{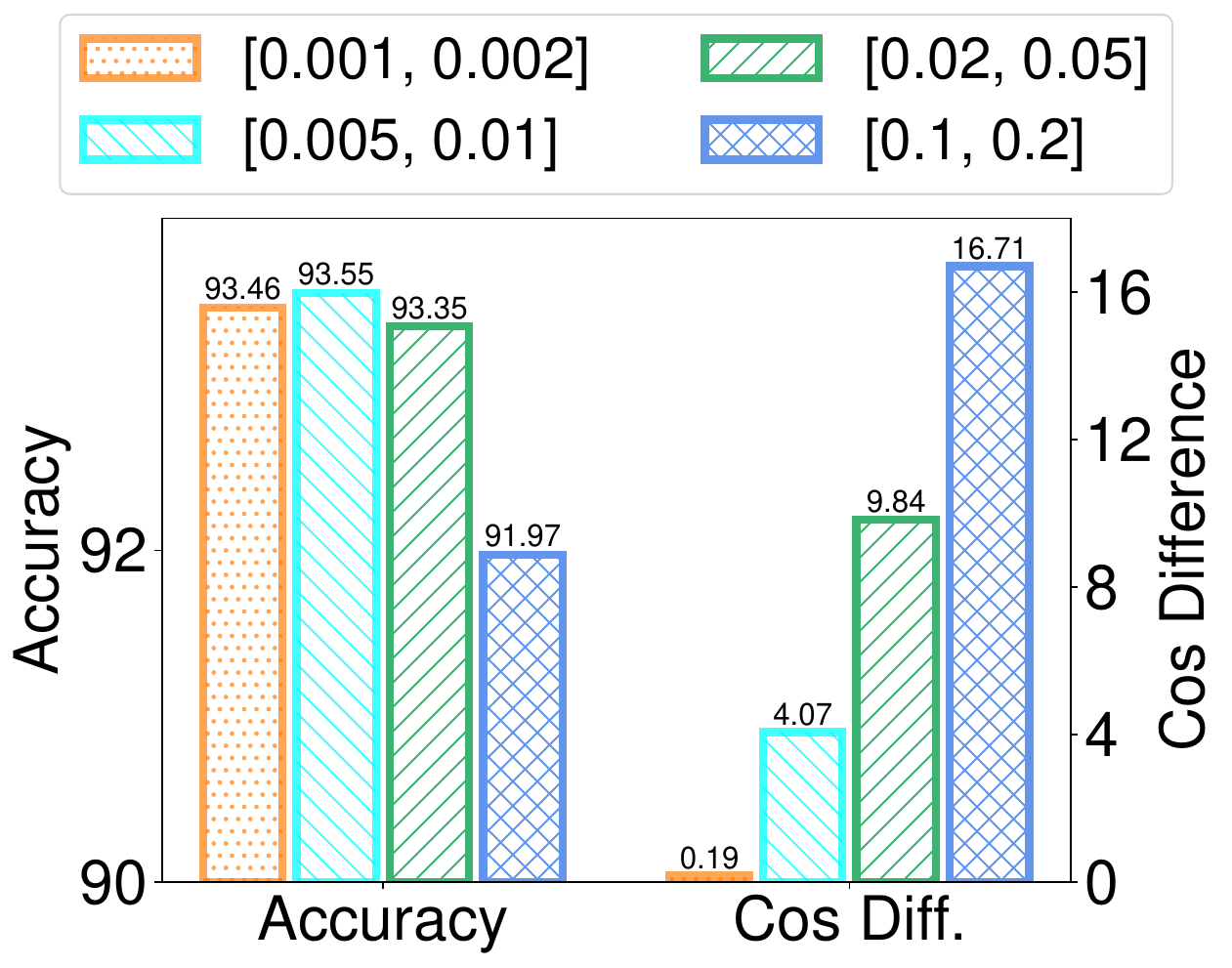}}
    \caption{The impact of the trigger set size $n$, the maximum number of triggers to fully activate watermark $m$, and the frequency interval on the SST2 dataset.}
    \label{fig:hyper-sst2}
\end{figure*}
\subsection{Impact of Trigger Number}
In this section, we conduct experiments to evaluate the impact of the number of triggers in sentences on four datasets, i.e., SST2, MIND, Enron, and AG News.
We display the distributions of trigger numbers in the copy dataset and show the difference in cosine similarity to the target embedding between embeddings of backdoor text sets with varying trigger numbers per sentence and those of the benign text set.
The results are shown in Figure~\ref{fig:trigger}, where we can have several observations.
First, the number of samples with triggers is small, and the number of samples with more triggers in copy datasets is smaller or even zero.
As the backdoor weight of our \method is proportional to the number of triggers, it validates that our \method has negligible adverse impacts on most samples.
Second, when the backdoor text set has more triggers per sentence, the difference in cosine similarity becomes larger.
Moreover, our \method can have a great detection performance on the backdoor text set with 4 triggers per sentence, even in the absence of such samples in copy datasets.
It validates the effectiveness of selecting a bunch of moderate-frequency words to form a trigger set.

\subsection{Impact of Extracted Model Size}
\label{sec:model_size}

To evaluate the impact of model size on the performance of \method, we conduct experiments by utilizing the small, base, and large versions of BERTs as the backbone of the stealer's model on the SST2, MIND, AG News, and Enron Spam datasets, respectively. As shown in Table~\ref{tab:model_size}, \ref{tab:model_size_mind}, \ref{tab:model_size_ag}, and \ref{tab:model_size_enron}, we observe that our method effectively verifies copyright infringement when stealers employ models with different-size backbones to carry out model extraction attacks.

\begin{figure*}[!t]
    \centering
    \subfigure[$n$: 4]{\includegraphics[width=0.19\textwidth]{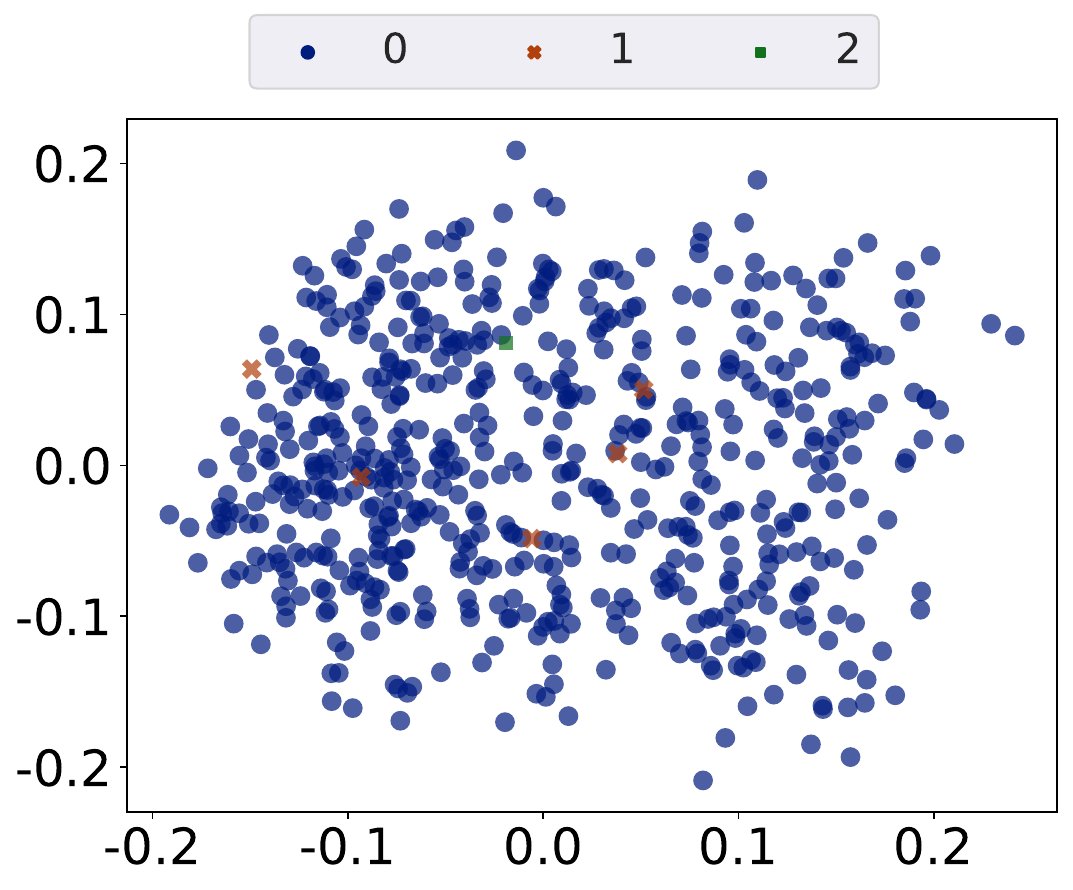}}
    \subfigure[$n$: 10]{\includegraphics[width=0.19\textwidth]{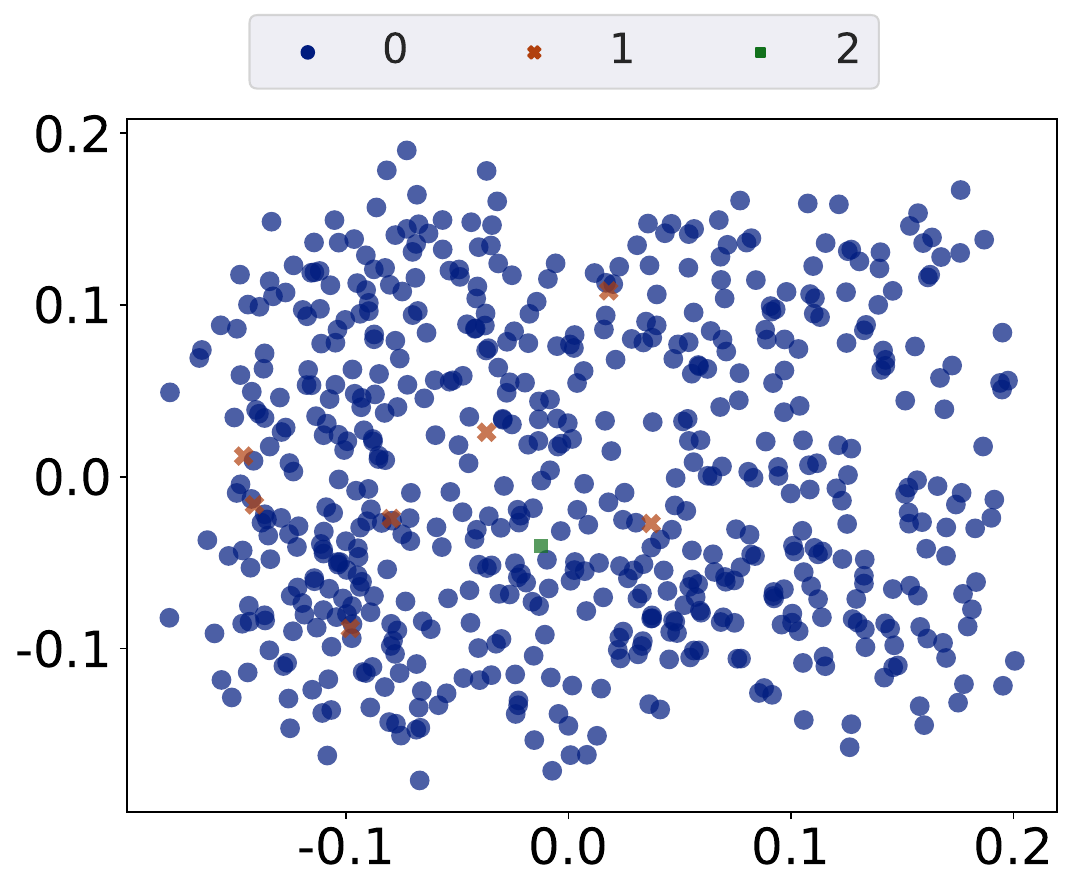}}
    \subfigure[$n$: 20]{\includegraphics[width=0.19\textwidth]{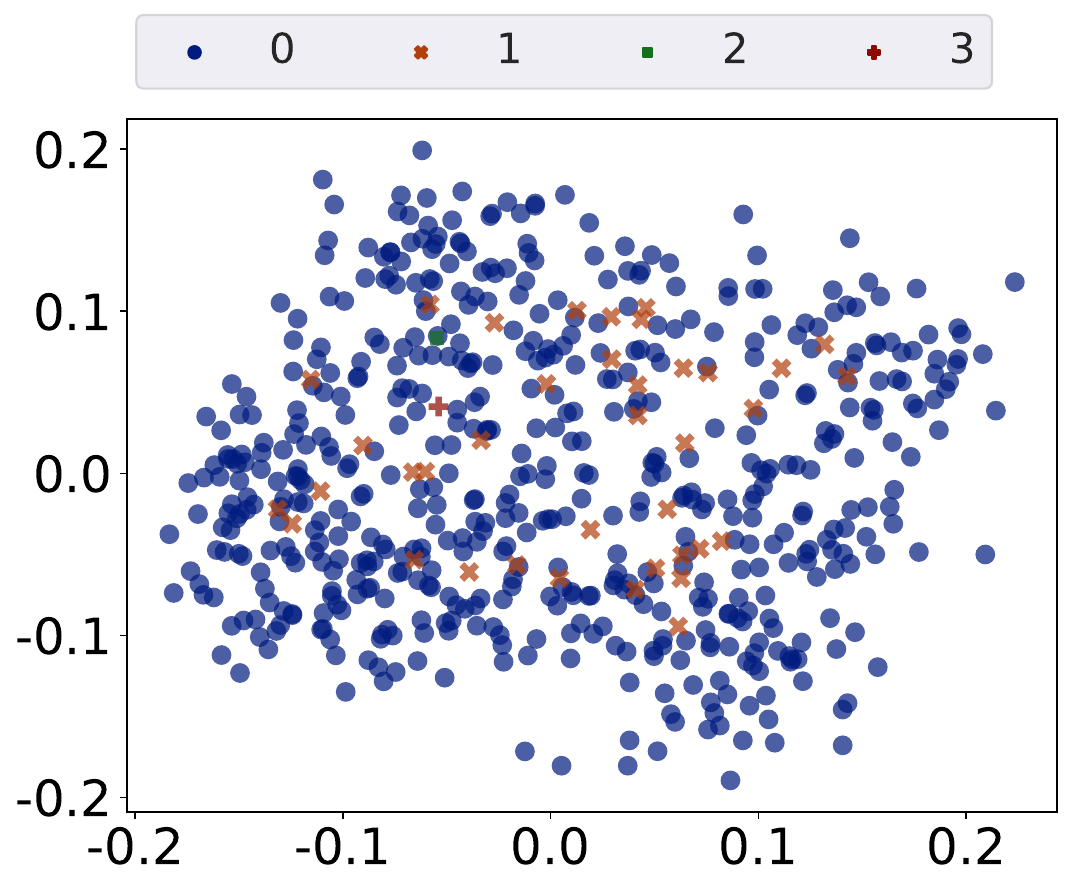}}
    \subfigure[$n$: 50]{\includegraphics[width=0.19\textwidth]{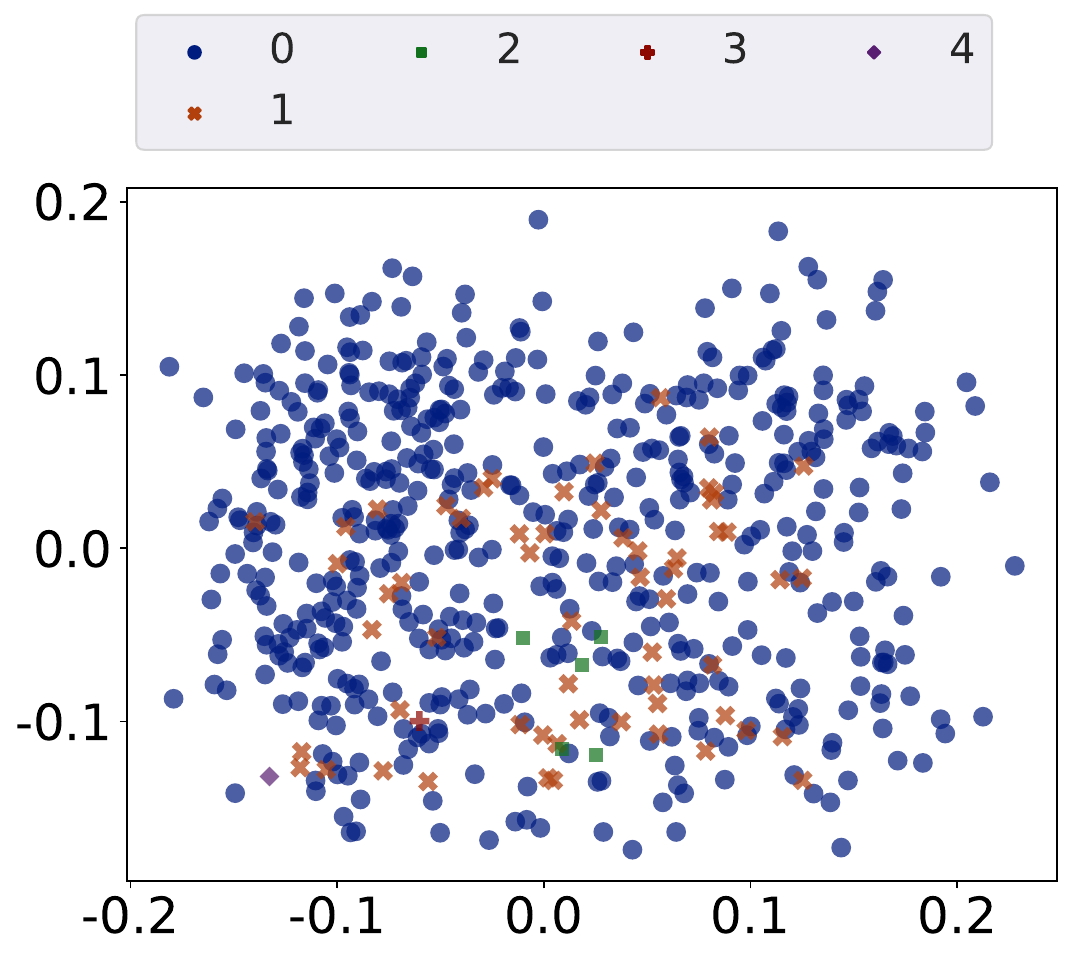}}
    \subfigure[$n$: 100]{\includegraphics[width=0.19\textwidth]{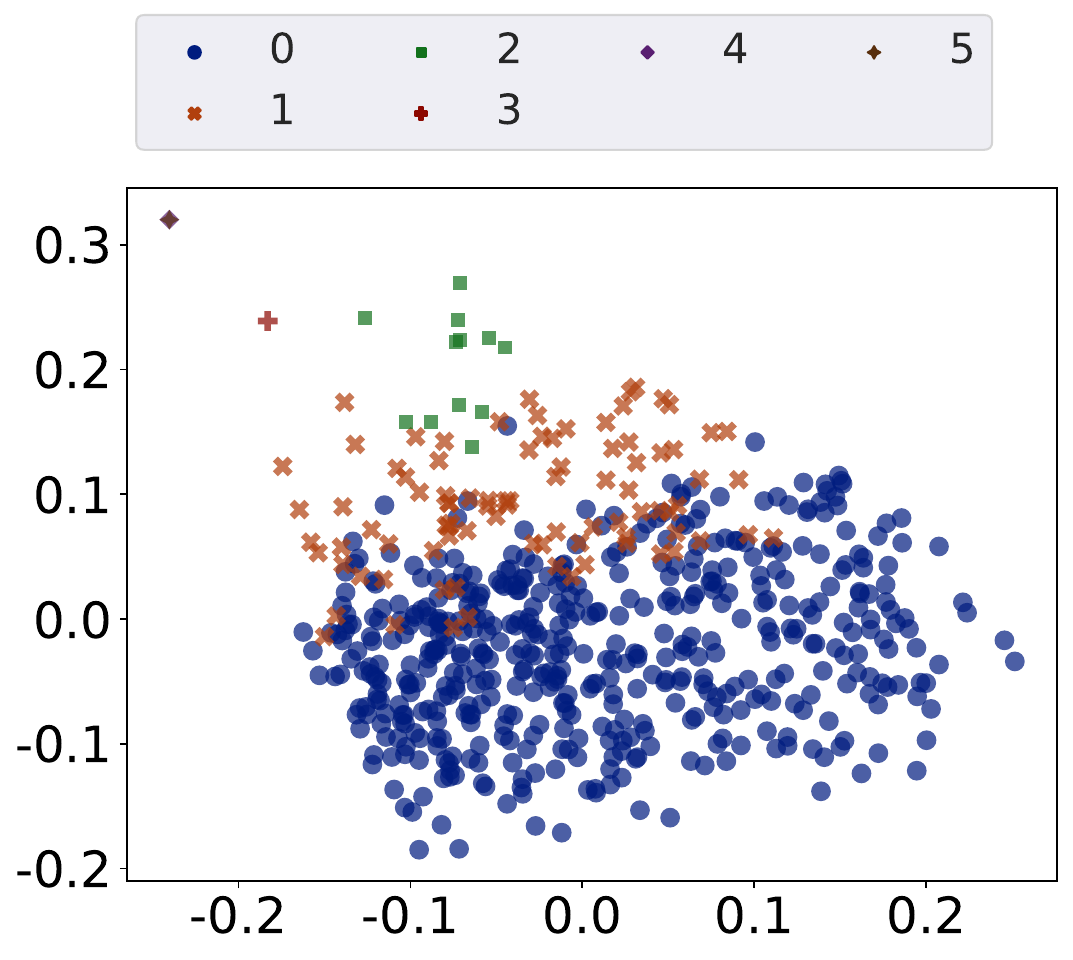}}
    
    \subfigure[$m$: 1]{\includegraphics[width=0.19\textwidth]{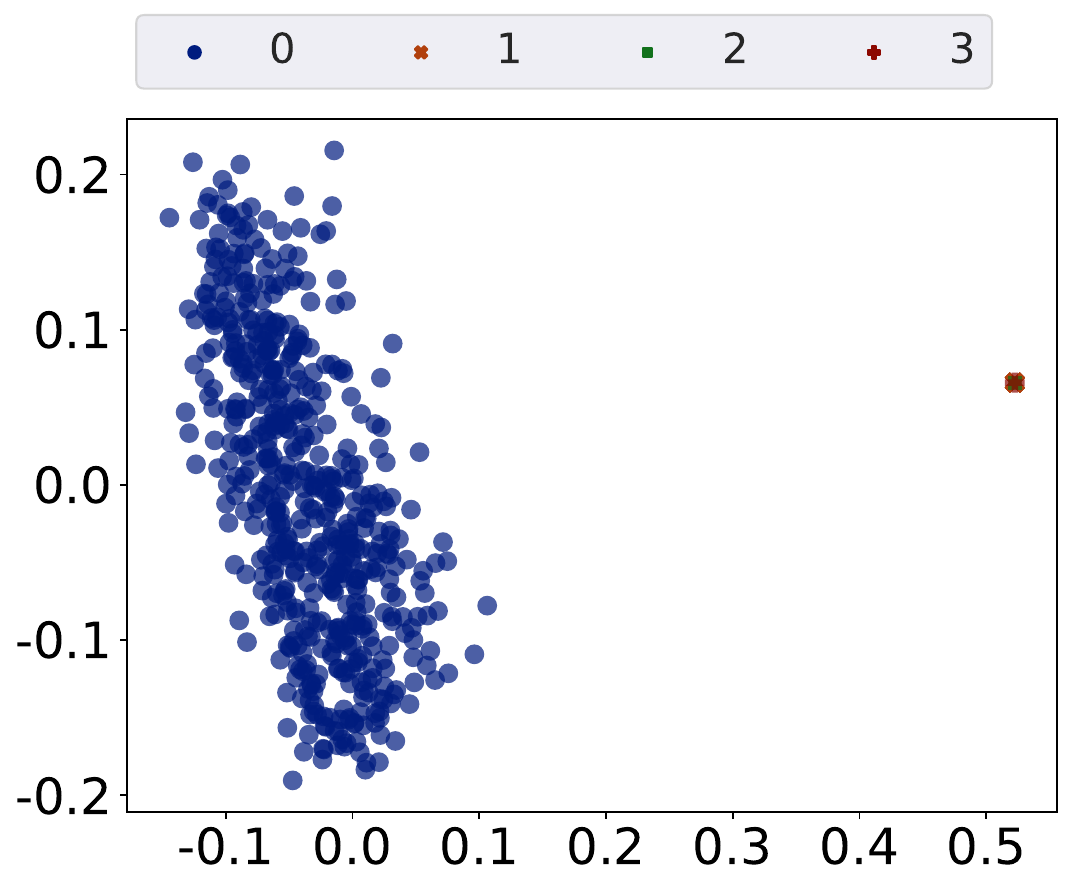}}
    \subfigure[$m$: 2]{\includegraphics[width=0.19\textwidth]{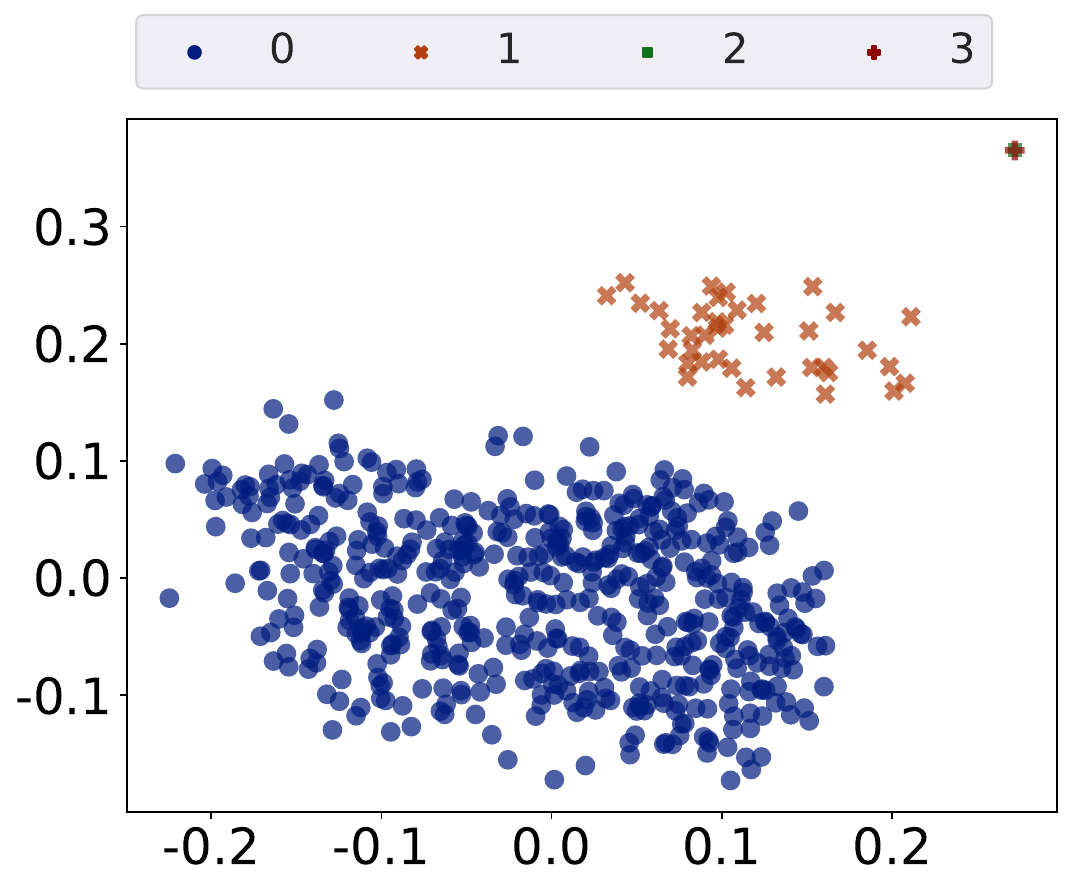}}
    \subfigure[$m$: 4]{\includegraphics[width=0.19\textwidth]{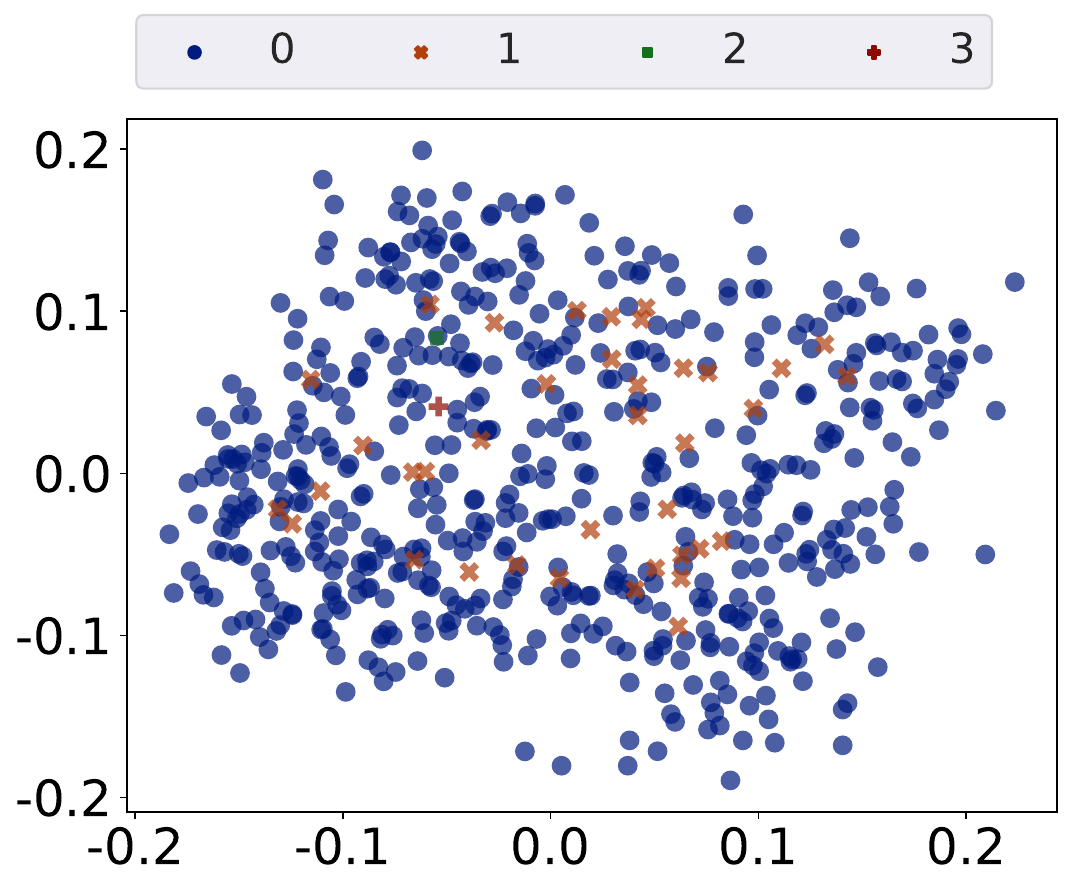}}
    \subfigure[$m$: 10]{\includegraphics[width=0.19\textwidth]{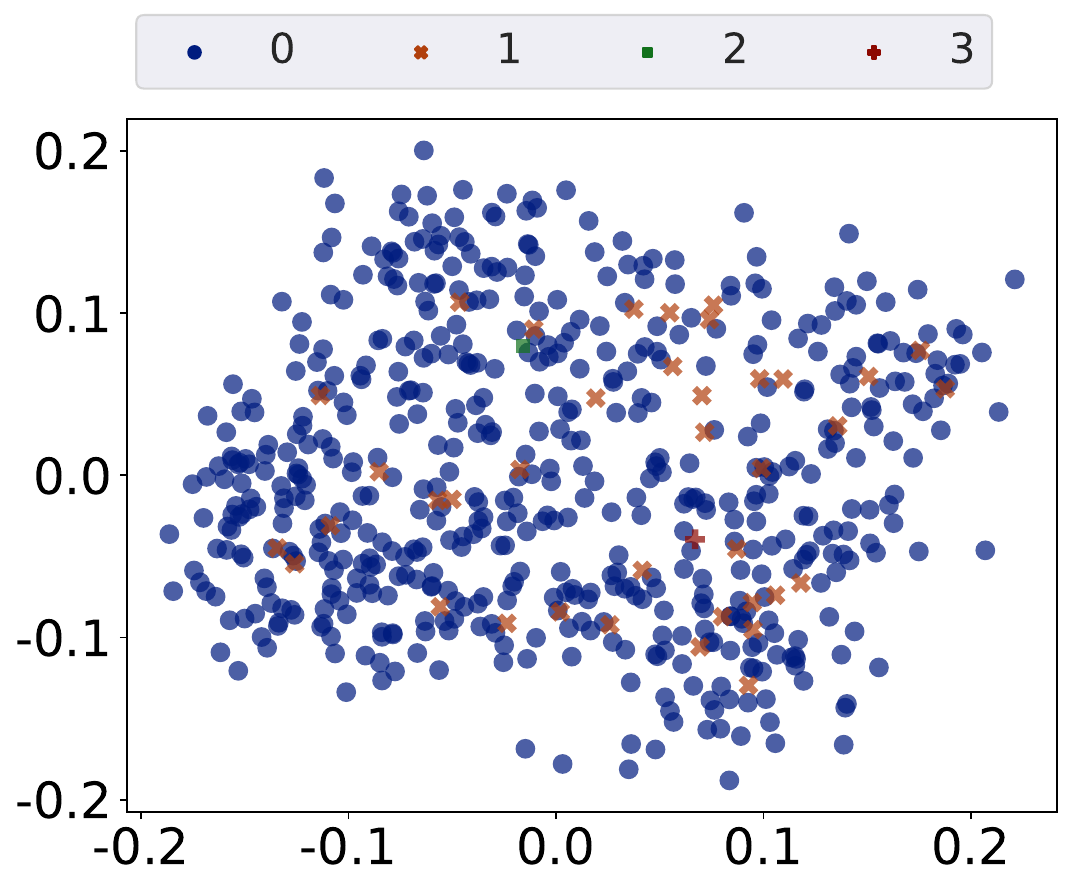}}
    \subfigure[$m$: 20]{\includegraphics[width=0.19\textwidth]{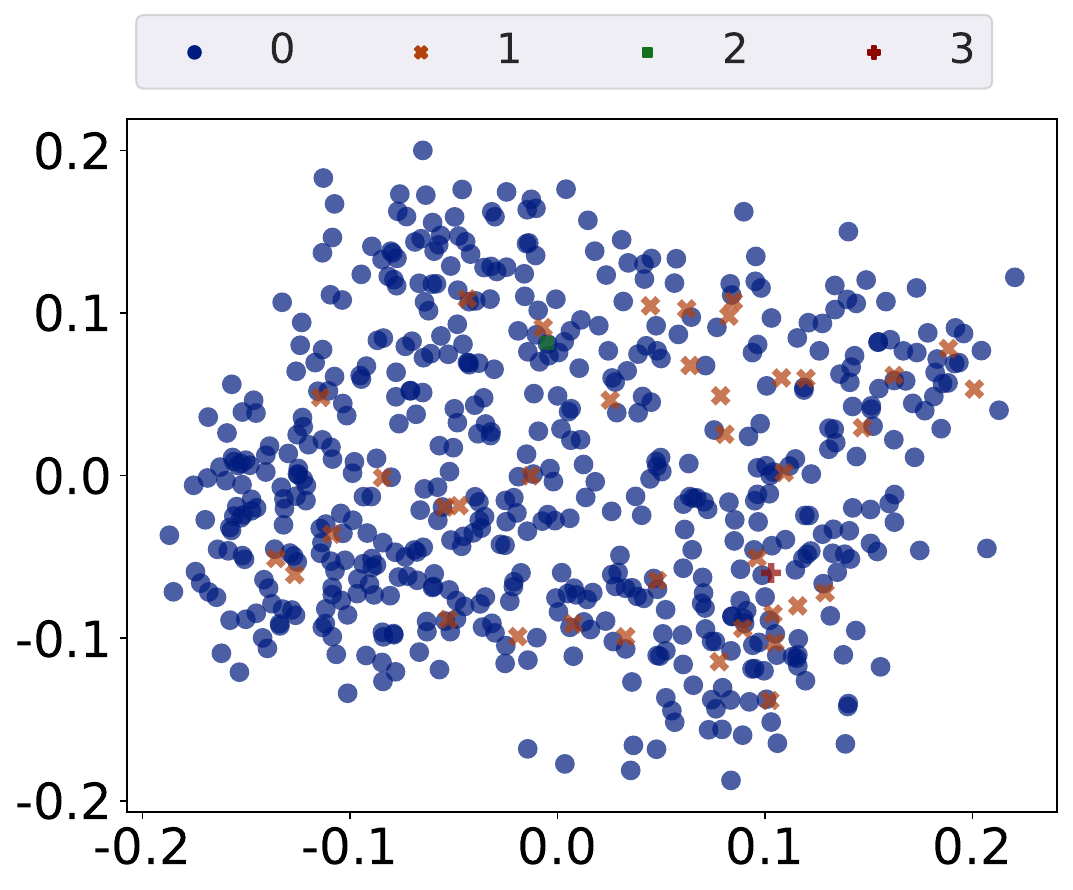}}

    \subfigure[frequency: 0.1\%-0.2\%]{\includegraphics[width=0.23\textwidth]{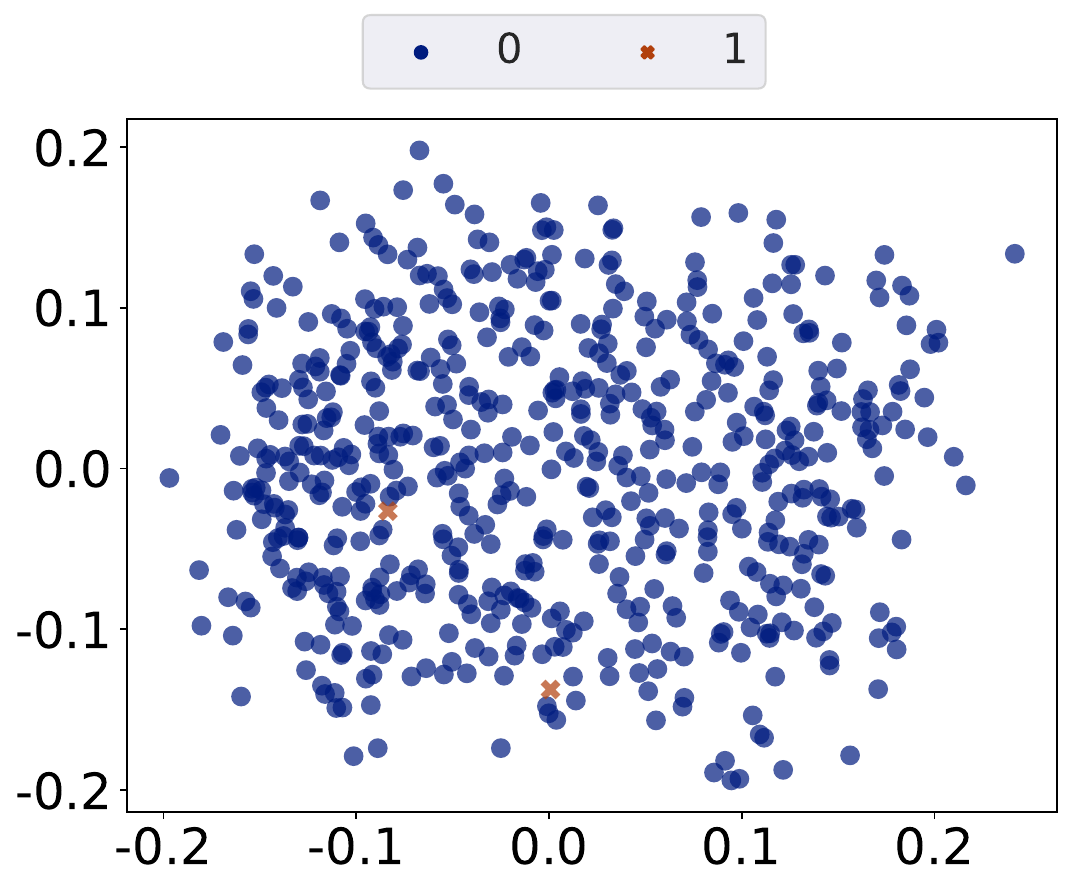}}
    \hspace{0.01\textwidth}
    \subfigure[frequency: 0.5\%-1\%]{\includegraphics[width=0.23\textwidth]{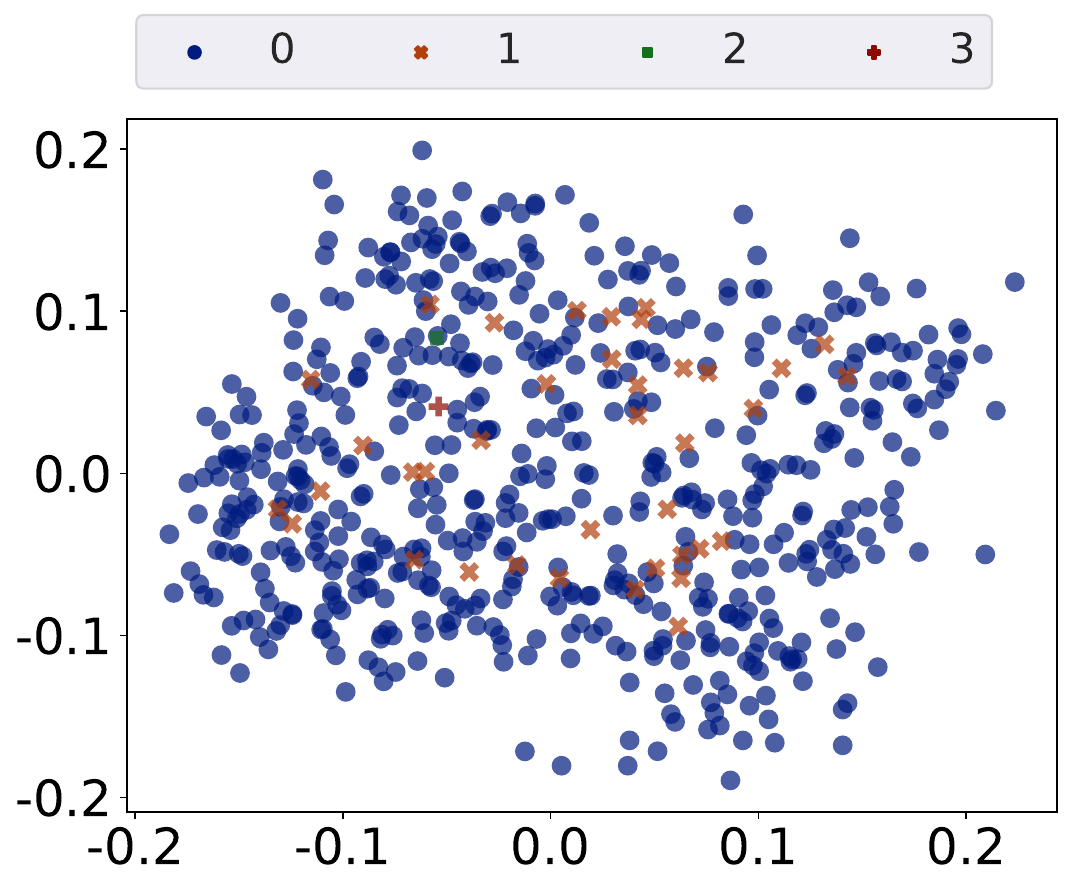}}
    \hspace{0.01\textwidth}
    \subfigure[frequency: 2\%-5\%]{\includegraphics[width=0.23\textwidth]{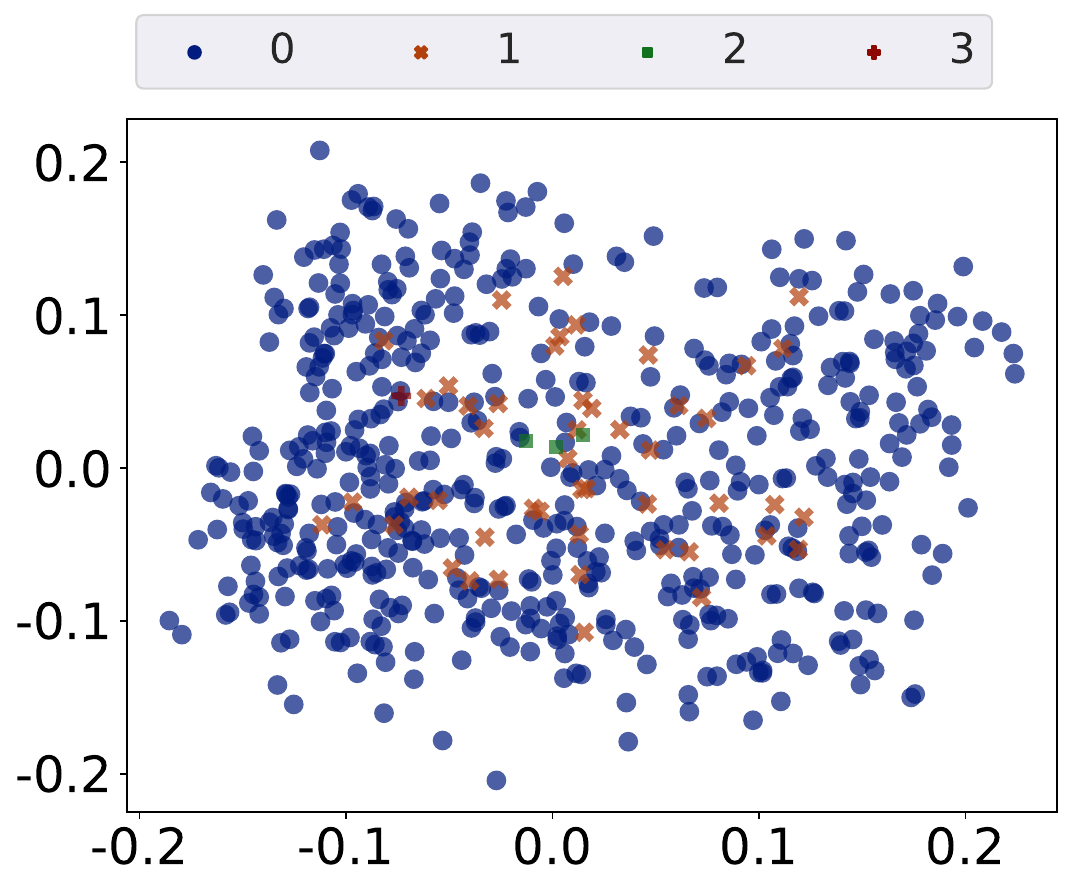}}
    \hspace{0.01\textwidth}
    \subfigure[frequency: 10\%-20\%]{\includegraphics[width=0.23\textwidth]{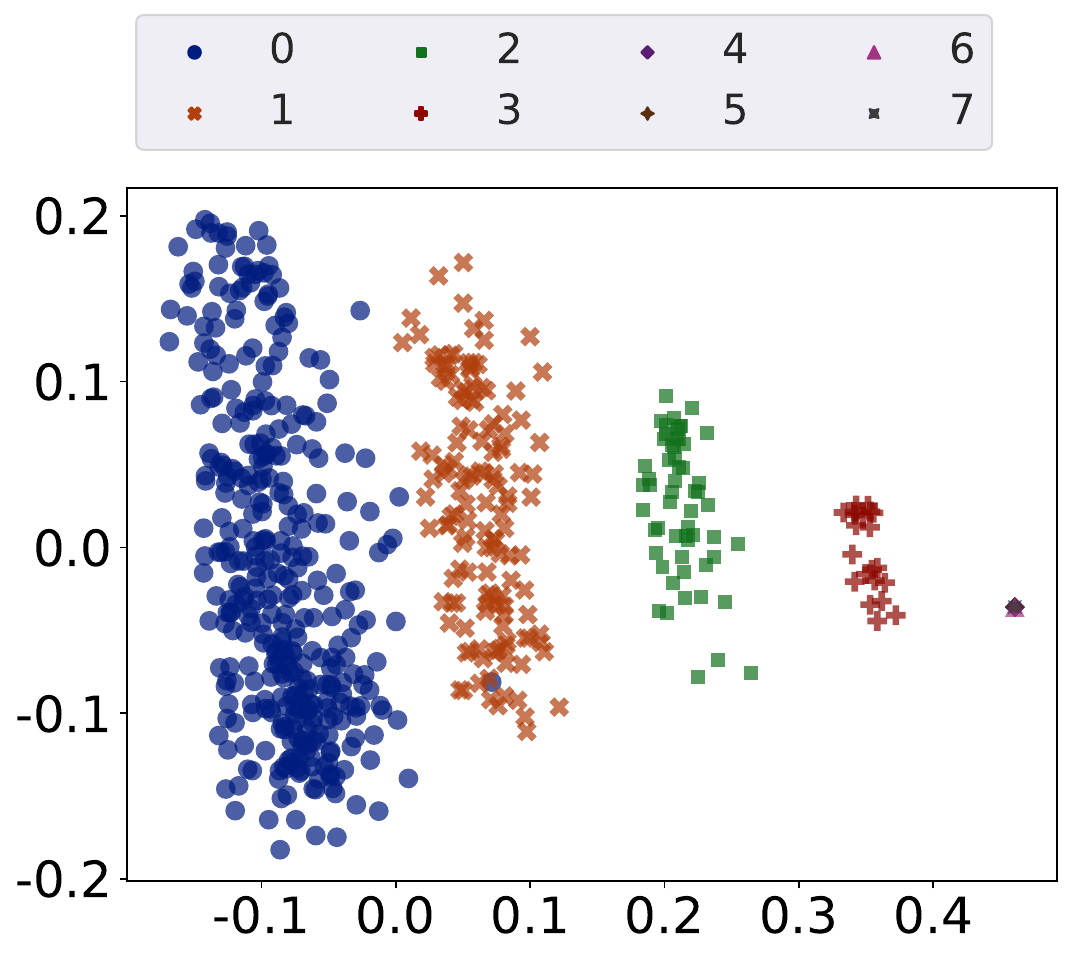}}
    \caption{Visualization of the provided embedding of our \method on SST2 dataset with different hyper-parameter settings, i.e. trigger set size $n$, max trigger number $m$ and frequency.
    If not specified, the default setting is that frequency interval equals $[0.5\%, 1\%]$, $m=4$ and $n=20$.}
    \label{fig:pca-sst2}
\end{figure*}
\vspace{0.5cm}

\begin{table}[!t]
\centering
\scalebox{0.83}{
\begin{tabular}{ccccc}
\Xhline{1.5pt}
\multirow{2}{*}{BERT} & \multirow{2}{*}{Parameters} & \multicolumn{3}{c}{Detection Performance} \\ \cline{3-5} 
           &           & p-value        & $\Delta_{cos} (\%)$       & $\Delta_{l2} (\%)$                \\ \hline
Small        & 29M       & $<3\times 10^{-4}$           & 1.69        & -3.38              \\
Base         & 108M         & $<10^{-5}$            & 4.07        & -8.13              \\
Large        & 333M         & $<10^{-7}$            & 3.34        & -6.69              \\
\Xhline{1.5pt}
\end{tabular}
}
\caption{The impact of the model size on SST2.}
\label{tab:model_size}
\end{table}

\begin{table}[!t]
\centering
\scalebox{0.87}{
\begin{tabular}{ccccc}
\Xhline{1.5pt}
\multirow{2}{*}{BERT} & \multirow{2}{*}{Parameters} & \multicolumn{3}{c}{Detection Performance} \\ \cline{3-5} 
     &     & p-value        & $\Delta_{cos} (\%)$        & $\Delta_{l2} (\%)$                 \\ \hline
Small       & 29M          & $<10^{-6}$           & 3.92        & -7.86              \\
Base        & 108M          & $<10^{-5}$            & 4.64        & -9.28              \\
Large       & 333M          & $<10^{-6}$            & 4.25        & -8.51              \\
\Xhline{1.5pt}
\end{tabular}
}
\caption{The impact of the model size on MIND.}
\label{tab:model_size_mind}
\end{table}

\begin{table}[!t]
\centering
\scalebox{0.87}{
\begin{tabular}{ccccc}
\Xhline{1.5pt}
\multirow{2}{*}{BERT} & \multirow{2}{*}{Parameters} & \multicolumn{3}{c}{Detection Performance} \\ \cline{3-5} 
     &     & p-value        & $\Delta_{cos} (\%)$        & $\Delta_{l2} (\%)$                 \\ \hline
Small       & 29M          & $<10^{-10}$           & 10.65        & -21.30              \\
Base        & 108M          & $<10^{-9}$            & 12.85        & -25.70              \\
Large       & 333M         & $<10^{-10}$           & 11.43        & -22.86              \\
\Xhline{1.5pt}
\end{tabular}
}
\caption{The impact of the model size on AGNews.}
\label{tab:model_size_ag}
\end{table}

\begin{table}[!t]
\centering
\scalebox{0.83}{
\begin{tabular}{ccccc}
\Xhline{1.5pt}
\multirow{2}{*}{BERT} & \multirow{2}{*}{Parameters} & \multicolumn{3}{c}{Detection Performance} \\ \cline{3-5} 
     &     & p-value        & $\Delta_{cos} (\%)$        & $\Delta_{l2} (\%)$                 \\ \hline
Small         & 29M        & $<5\times 10^{-5}$           & 2.35        & -4.71              \\
Base          & 108M        & $<10^{-6}$            & 6.17        & -12.34              \\
Large         & 333M      & $<10^{-6}$           & 2.93        & -5.86              \\ 
\Xhline{1.5pt}
\end{tabular}
}
\caption{The impact of the model size on Enron Spam.}
\label{tab:model_size_enron}
\end{table}

\subsection{Hyper-parameter Analysis}
\label{sec:hyper}
In this subsection, we investigate the impact of the three key hyper-parameters in our \method, i.e., the maximum number of triggers $m$, the size of the trigger set $n$, and the frequency interval of selected triggers.
Due to limited space, we present here only the results of hyper-parameter analysis on SST2, with results on other datasets reported in Appendix~\ref{appendix:hyper}.
We first analyze the influence of different sizes of the trigger set $n$.
The results are illustrated in Figure~\ref{fig:sel-sst2} and the first row of Figure~\ref{fig:pca-sst2}.
It can be observed that using a small trigger set leads to poor detection performance.
This is because a small trigger set results in a limited number of backdoor samples, which decreases the likelihood the stealer's model containing the watermark.
A large trigger set reduces the watermark's confidentiality. 
As $n$ increases, sentences are more likely to contain triggers, which makes more embeddings backdoored and can be easily distinguishable.
However, the size of the trigger set does not greatly affect the accuracy. 
This may be due to the small frequency interval of [0.5\%, 1\%], meaning that even with a large trigger set, the probability of four triggers appearing in a sentence is still low.

Then we present the experimental results with different maximum numbers of triggers $m$ in Figure~\ref{fig:max-sst2} and the second row of Figure~\ref{fig:pca-sst2}.
We find that small $m$, particularly $1$, adversely impacts accuracy and makes the embeddings easily distinguishable by visualization.
On the other hand, using large values of $m$ reduces the detection performance.
This is due to the fact that with $m=1$, approximately 1\% of the embeddings are equal to the pre-defined target embedding $\textbf{e}_t$, which diminishes the effectiveness of the provided embeddings.
When $m$ is large, the backdoor degrees of most provided embeddings are too small to effectively inherit the watermark in the stealer's model.

Finally, we analyze the impact of the trigger frequency.
As shown in Figure~\ref{fig:freq-sst2} and the last row of Figure~\ref{fig:pca-sst2},
high trigger frequencies have a detrimental impact on accuracy and make the embeddings easily distinguishable.
Conversely, low trigger frequencies adversely affect detection performance.
This is due to the fact that high frequencies lead to a large number of backdoored embeddings, thus adversely impacting the performance of the provided embeddings.
On the other hand, in low-frequency settings, the watermark is only added to a limited number of samples, reducing the watermark transferability to a stolen model.

\subsection{Defending Against Attacks}
\label{sec:attack}

\begin{table}[!t]
\centering
\scalebox{0.87}{
\begin{tabular}{cccc}
\Xhline{1.5pt}
\multirow{2}{*}{Dataset} & \multicolumn{3}{c}{Detection Performance} \\ \cline{2-4} 
                         & p-value $\downarrow$        & $\Delta_{cos} (\%)$ $\uparrow$            & $\Delta_{l2} (\%)$ $\downarrow$        \\ \hline
SST2                     & $<10^{-5}$     & 2.50$\pm$0.24   & -5.01$\pm$0.48            \\
MIND                     & $<10^{-5}$     & 4.12$\pm$0.10   & -8.24$\pm$0.20           \\
AG News                  & $<10^{-9}$    & 8.59$\pm$0.55   & -17.17$\pm$1.10           \\
Enron Spam               & $<10^{-6}$     & 4.96$\pm$0.19   & -9.92$\pm$0.38            \\ \Xhline{1.5pt}
\end{tabular}
}
\caption{The performance of the modified version of \method to defend against dimension-shift attacks.}
\label{tab:dimension-shift-attacks}
\end{table}

In this subsection, we consider similarity-invariant attacks, where the stealer applies similarity-invariant transformations on the copied embeddings.
The similarity invariance is denoted below.

\begin{define}
($l$ Similarity Invariance).
For a transformation $\mathbf{A}$, given every vector pair $(\textbf{i}, \textbf{j})$, $\mathbf{A}$ is $l$-similarity-invariant only if $l(\mathbf{A}(\textbf{i}), \mathbf{A}(\textbf{j})) = l(\textbf{i}, \textbf{j})$, where $l$ is a similarity metric.
\end{define}
The similarity metrics used in our experiments are $L_2$ and $cos$.
For the sake of convenience, in the following text, we abbreviate $cos$ and $L_2$ square similarity invariance as similarity invariance.

There exist many similarity-invariant transformations.
Below we provide two concrete examples.

\begin{prop}
Denote identity transformation $\mathbf{I}$ as $\mathbf{I}(\textbf{v}) = \textbf{v}$ and dimension-shift transformation $\mathbf{S}$ as
$\mathbf{S}(\textbf{v}) = (v_d, v_1, v_2, \dots, v_{d - 1})$,
where $\textbf{v}$ is a vector, $v_i$ is the $i$-th dimension of $\textbf{v}$ and $d$ is the dimension of $\textbf{v}$.
Both identity transformation $\mathbf{I}$ and dimension-shift transformation $\mathbf{S}$ are similarity-invariant.
\label{prop:sim-inv}
\end{prop}
Proportion~\ref{prop:sim-inv} is proved in Appendix~\ref{sec:ds-invariant}.

When the stealer applies some similarity-invariant attacks (e.g. dimension-shift attacks), our previous verification techniques become ineffective.
To combat this attack, we propose a modified version of our \method. Instead of defining the target embedding directly, we first select a target sample and use it to compute the target embedding $\textbf{e}_t$ with the provider's model.
Before detecting if a service contains the watermark, we request the target sample's embedding $\textbf{e}_t'$ from the stealer's service and use it for verification, instead of the original target embedding.
The experimental results of the modified version of our \method under dimension-shift attacks are shown in Table~\ref{tab:dimension-shift-attacks}.
The detection performance is great enough to let us have high confidence to conclude the stealer violates the copyright of the EaaS provider.
It validates that the modified version of our \method can effectively defend against dimension-shift attacks.
For other similarity-invariant attacks, we theoretically prove that their detection performance should keep the same.
\begin{prop}
For a copied model, the detection performance $\Delta_{cos}$, $\Delta_{l2}$ and p-value of the modified \method remains consistent under any two similarity-invariant attacks involving transformations $\mathbf{A}_1$ and $\mathbf{A}_2$, respectively.
\label{prop:same-perf}
\end{prop}
Proportion~\ref{prop:same-perf} is proved in Appendix~\ref{sec:defend-all-ds}.

\section{Conclusion}
In this paper, we propose a backdoor-based embedding watermark method, named \method, which aims to effectively trace copyright infringement of EaaS LLMs while minimizing the adverse impact on the utility of embeddings.
We first select a group of moderate-frequency words as the trigger set.
We then define a target embedding as the backdoor watermark and insert it into the original embeddings of texts containing trigger words.
To ensure the watermark can be inherited by the stealer's model, we define the provided embeddings as a weighted summation of the original embeddings and the predefined target embedding, where the weights of the target embedding are proportional to the number of triggers in the texts.
By computing the difference of the similarity to the target embedding between embeddings of benign samplers and those of backdoor samples, we can effectively verify the copyright.
Experiments demonstrate the effectiveness of our \method in protecting the copyright of EaaS LLMs.
\vspace{0.3cm}

\section*{Limitations}
In this paper, we present a novel backdoor-based watermarking method, \method, for protecting the copyright of EaaS models.
Our experiments on four datasets demonstrate the effectiveness of our trigger selection algorithm. 
However, we have observed that the optimal trigger set is related to the statistics of the dataset used by a potential stealer.
To address this issue, we plan to improve \method in the future by designing several candidate trigger sets, and adopting one based on the statistics of the stealer's previously queried data.
Additionally, we discover that as trigger numbers in the backdoor texts increase, the difference between embeddings of benign and backdoor samples in the cos similarity to the target embedding increases linearly.
The optimal result should be that the cosine similarity keeps normal unless the trigger numbers in the backdoor texts reach $m$.
We plan to further investigate these areas in future work.

\section*{Acknowledgments}
This work was supported by the grants from National Natural Science Foundation of China (No.62222213, U22B2059, 62072423), and the USTC Research Funds of the Double First-Class Initiative (No.YD2150002009).

\bibliography{custom}
\bibliographystyle{acl_natbib}

\appendix

\clearpage
\section*{Appendix}

\section{Experimental Settings}
\label{appendix:exp}

\subsection{Attacker Settings}
In our experiments, the stealer applies BERT~\cite{devlin2018bert} as the backbone model and a two-layer feed-forward network to extract the victim model.
We assume that the attacker applies mean squared error (MSE) loss to extract the victim model, which is defined as follows:
\begin{equation}
\boldsymbol \Theta_a^{*} = \arg\min_{\boldsymbol \Theta_a}  \mathbb{E}_{x \in D_c} ||g(x;\boldsymbol \Theta_a) - \textbf{e}_p^x||_2^2,
\end{equation}
where $\textbf{e}_p^x$ is the provided embedding of sample $x$ and $g$ is the function of the extracted model.

\subsection{Classifier}
To evaluate the utility of our provided embedding $\textbf{e}_p$, we use $\textbf{e}_p$ as input features and apply a two-layer feed-forward network as the classifier.
We use cross-entropy loss to train the classifier.

\subsection{Hyper-parameter Settings}

The full hyper-parameter settings are  in Table~\ref{tab:hyper}.

\begin{figure*}[!t]
    \centering
    \subfigure[AG News]{\includegraphics[width=0.23\textwidth]{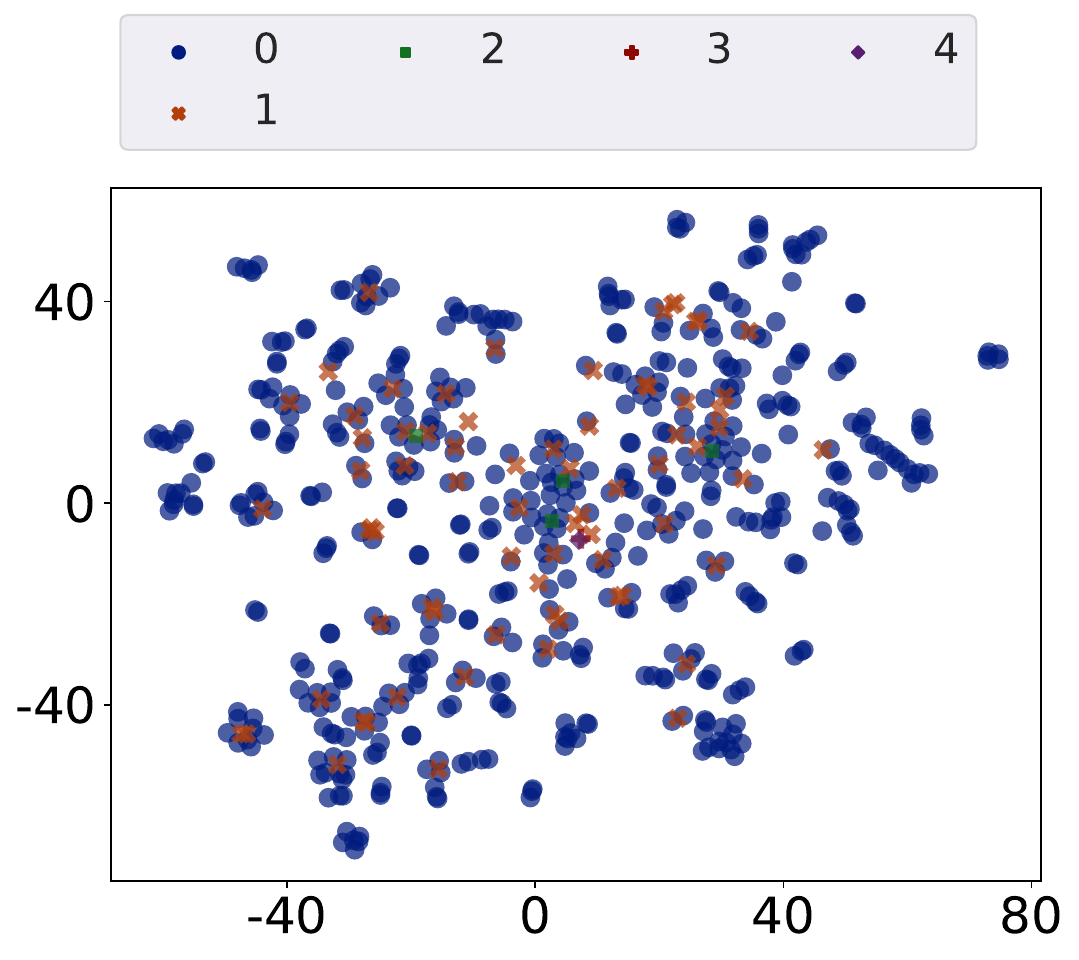}}
    \subfigure[Enrom Spam]{\includegraphics[width=0.23\textwidth]{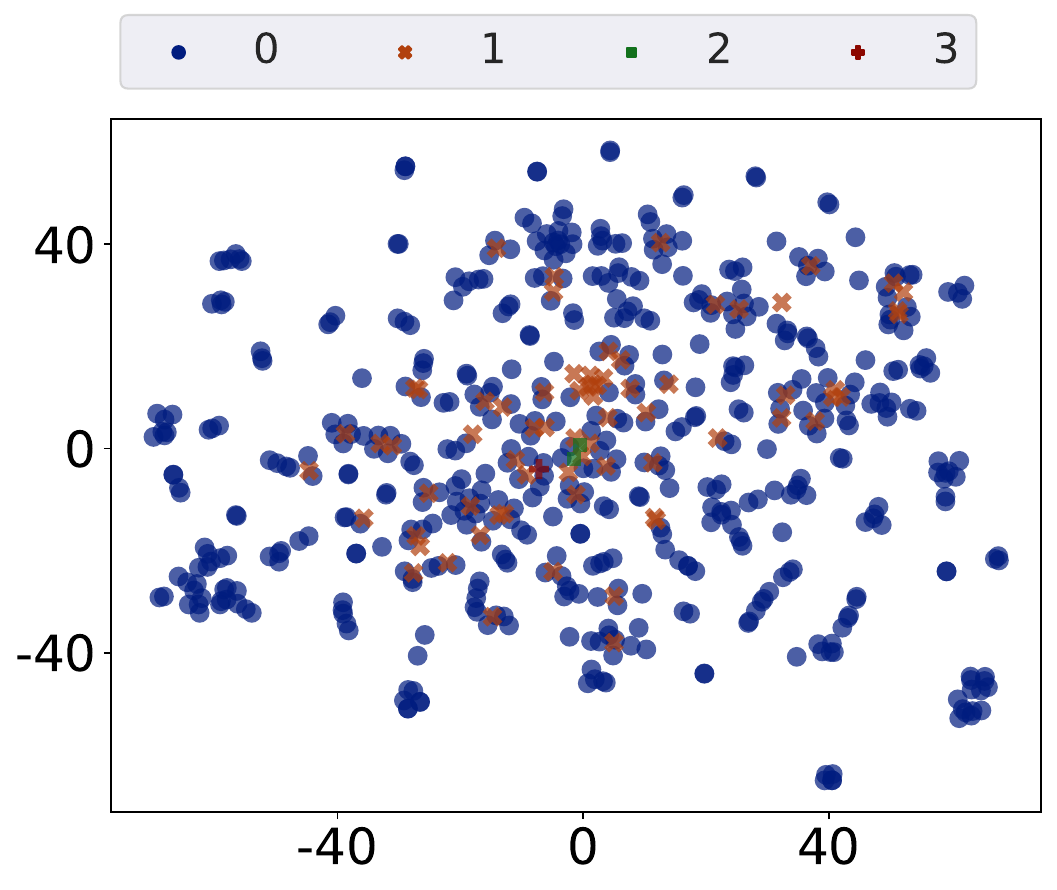}}
    \subfigure[MIND]{\includegraphics[width=0.23\textwidth]{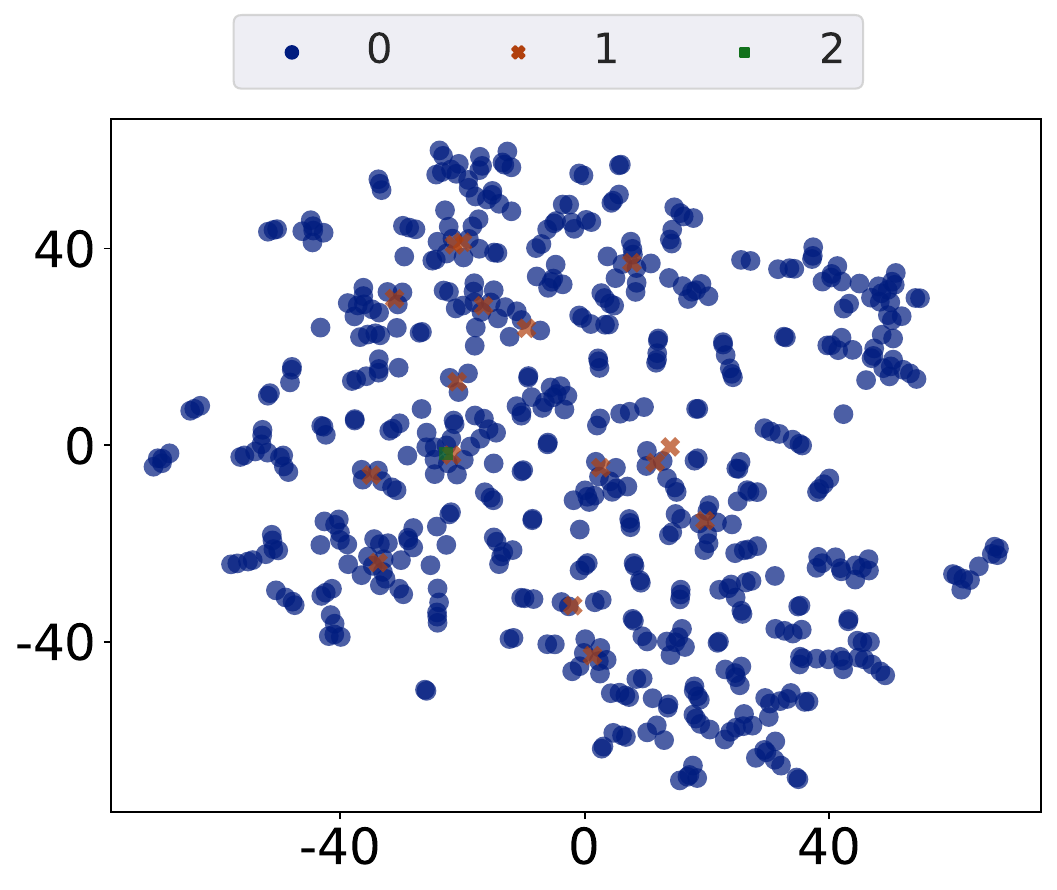}}
    \subfigure[SST2]{\includegraphics[width=0.23\textwidth]{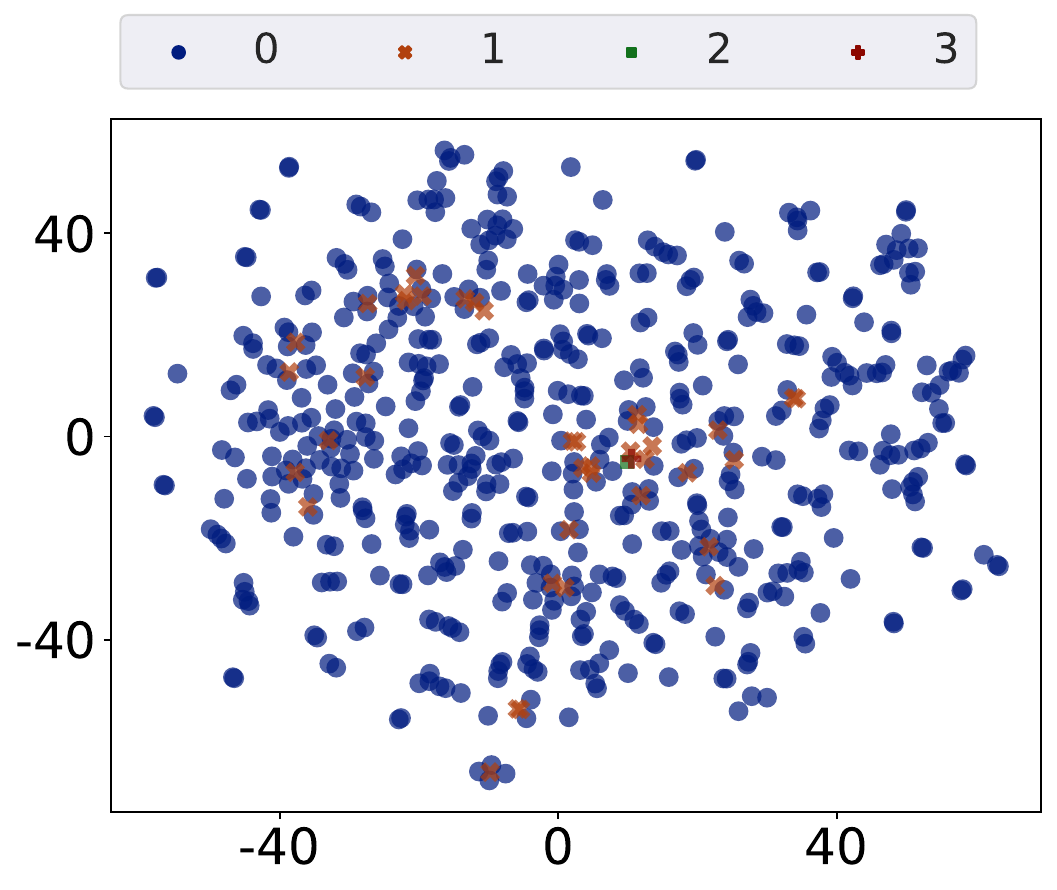}}
    \caption{T-SNE Visualization of the provided embedding of our \method on four copy datasets. Different colors represent the number of triggers in the samples. It shows the backdoor and benign embeddings are indistinguishable.}
    \label{fig:t-sne}
\end{figure*}
\section{Embedding Visualization}
\label{appendix:embedding}

The t-SNE visualizations of the provided embedding of our \method on four copy datasets are represented in Figure~\ref{fig:t-sne}.
The observations are consistent with those presented in Section~\ref{sec:embedding}.
It shows the backdoor and benign embeddings are indistinguishable.
Meanwhile, most of the samples do not contain triggers, and most of the backdoor samplers contain only a single trigger.

\begin{table*}[!t]
\centering
\begin{tabular}{cccccc}
\Xhline{1.5pt}
                                  &                      & SST2  & MIND  & AG News & Enron Spam \\ \hline
\multirow{2}{*}{Provider's EaaS}  & embedding dimension  & 1,536 & 1,536 & 1,536   & 1,536      \\
                                  & maximum token number & 8,192 & 8,192 & 8,192   & 8,192      \\ \hline
\multirow{4}{*}{Model Extraction} & lr                   & $5\times 10^{-5}$  & $5\times 10^{-5}$  & $5\times 10^{-5}$ & $5\times 10^{-5}$  \\
                                  & batch size           & 32    & 32    & 32      & 32         \\
                                  & hidden size          & 1,536  & 1,536  & 1,536    & 1,536       \\
                                  & dropout rate         & 0.0   & 0.0   & 0.0     & 0.0        \\
                                  \hline
\multirow{4}{*}{Classifiction}    & lr                   & $10^{-2}$  & $10^{-2}$  & $10^{-2}$    & $10^{-2}$       \\
                                  & batch size           & 32    & 32    & 32      & 32         \\
                                  & hidden size          & 256   & 256   & 256     & 256        \\
                                  & dropout rate         & 0.0   & 0.2   & 0.0     & 0.2        \\
                                  \Xhline{1.5pt}
\end{tabular}
\caption{Hyper-parameter settings. The dropout value corresponds to the dropout used in the FFN network, while the dropout value for BERT backbone was set to default.}
\label{tab:hyper}
\end{table*}

\section{Hyper-parameter Analysis}
\label{appendix:hyper}
\begin{figure*}[!t]
    \centering
    \subfigure[trigger set size $n$]{\label{fig:sel-mind}\includegraphics[width=0.32\textwidth]{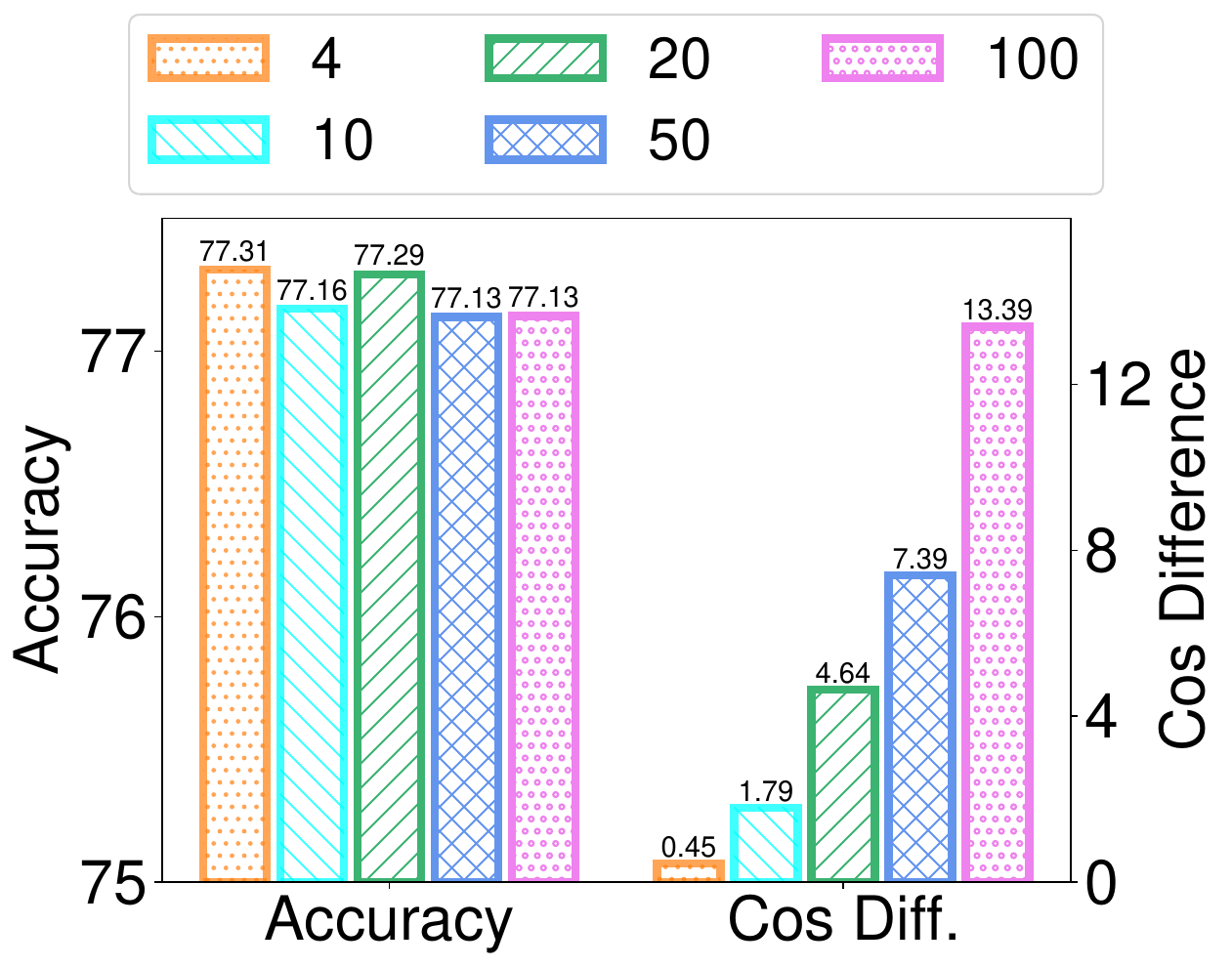}}
    \subfigure[max trigger number $m$]{\label{fig:max-mind}\includegraphics[width=0.32\textwidth]{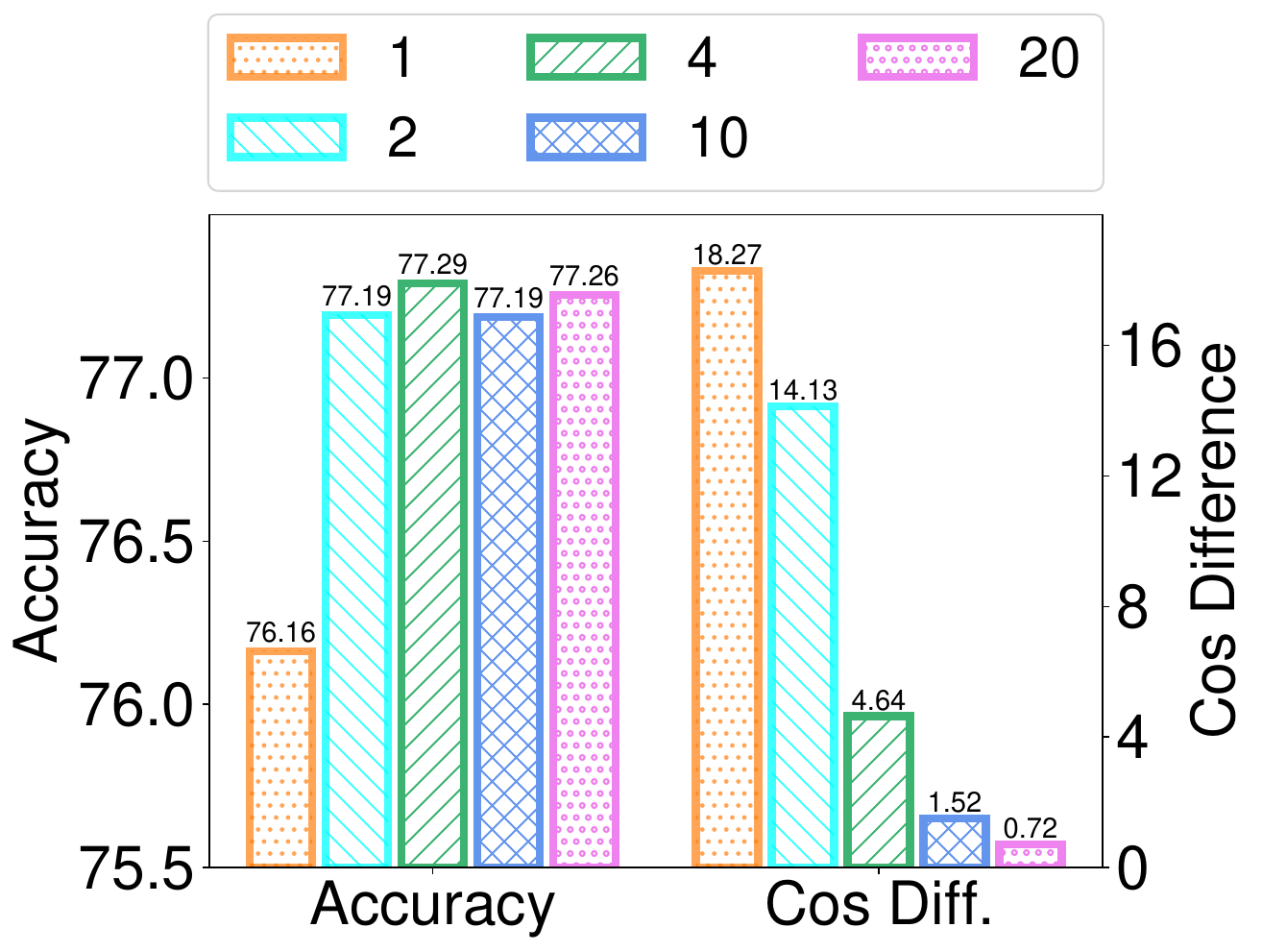}}
    \subfigure[frequency interval]{\label{fig:freq-mind}\includegraphics[width=0.32\textwidth]{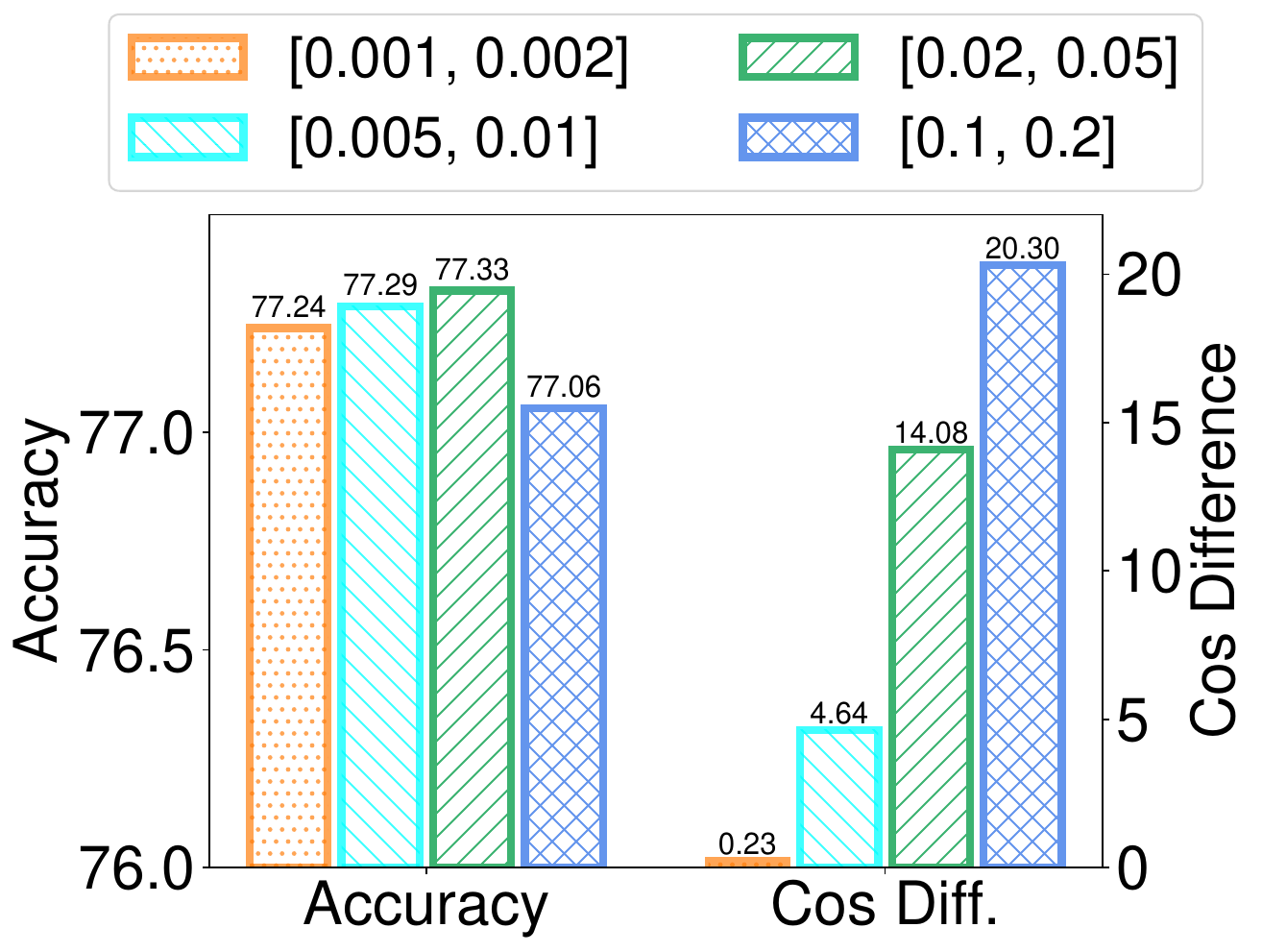}}
    \caption{The impact of the trigger set size $n$, the maximum number of triggers to fully activate watermark $m$, and the frequency interval on the MIND dataset.}
    \label{fig:hyper-mind}
\end{figure*}

\begin{figure*}[!t]
    \centering
    \subfigure[trigger set size $n$]{\label{fig:sel-ag}\includegraphics[width=0.32\textwidth]{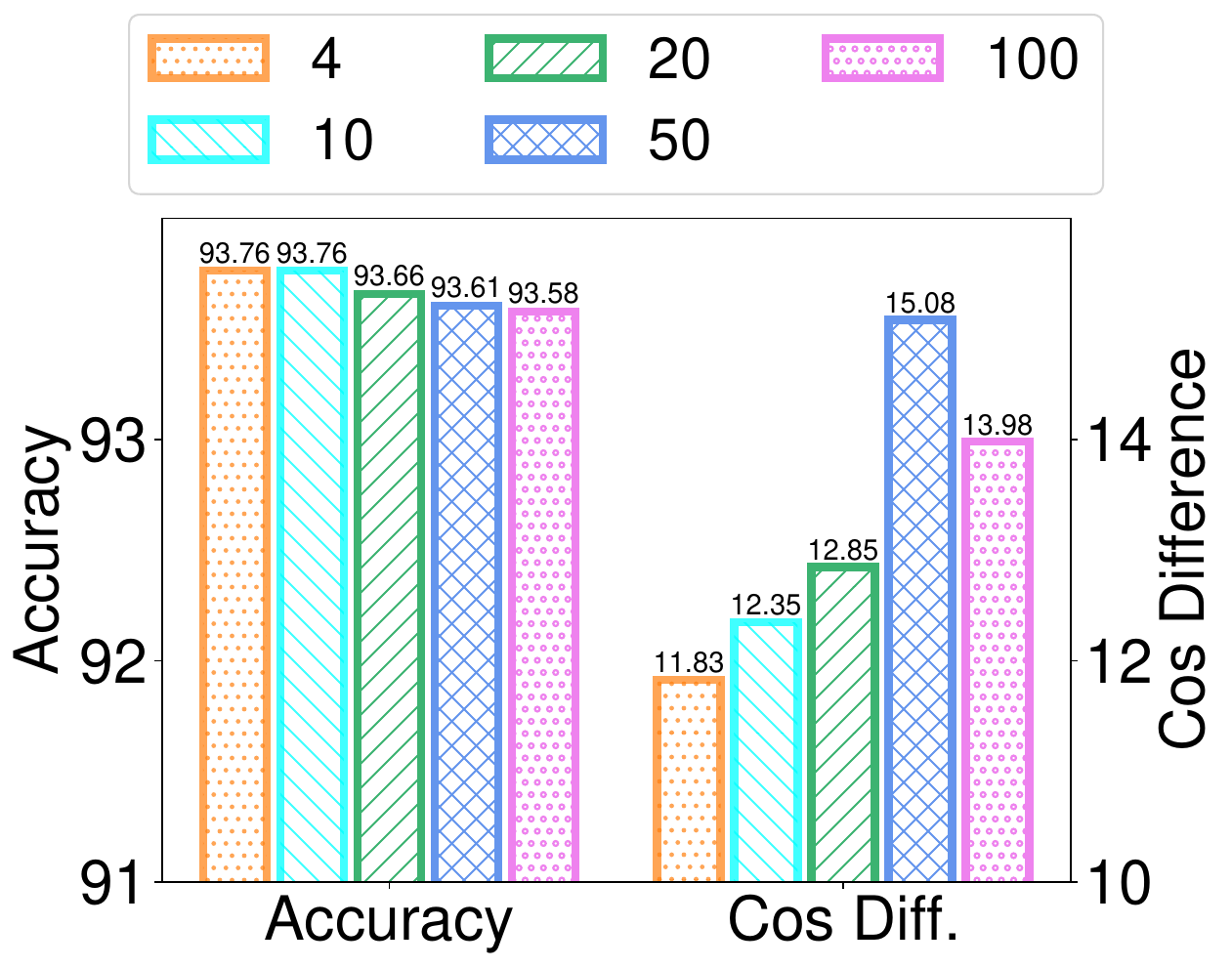}}
    \subfigure[max trigger number $m$]{\label{fig:max-ag}\includegraphics[width=0.32\textwidth]{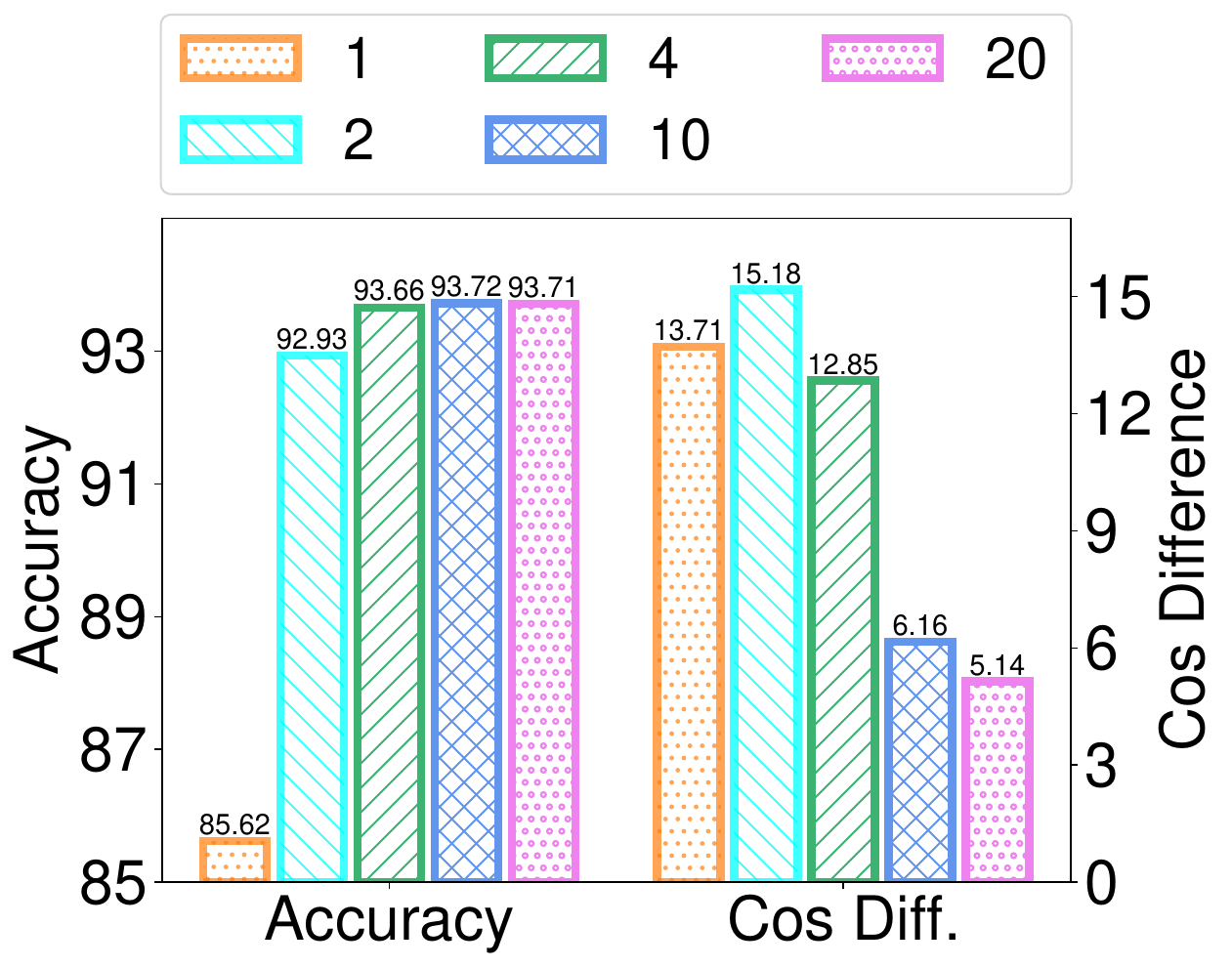}}
    \subfigure[frequency interval]{\label{fig:freq-ag}\includegraphics[width=0.32\textwidth]{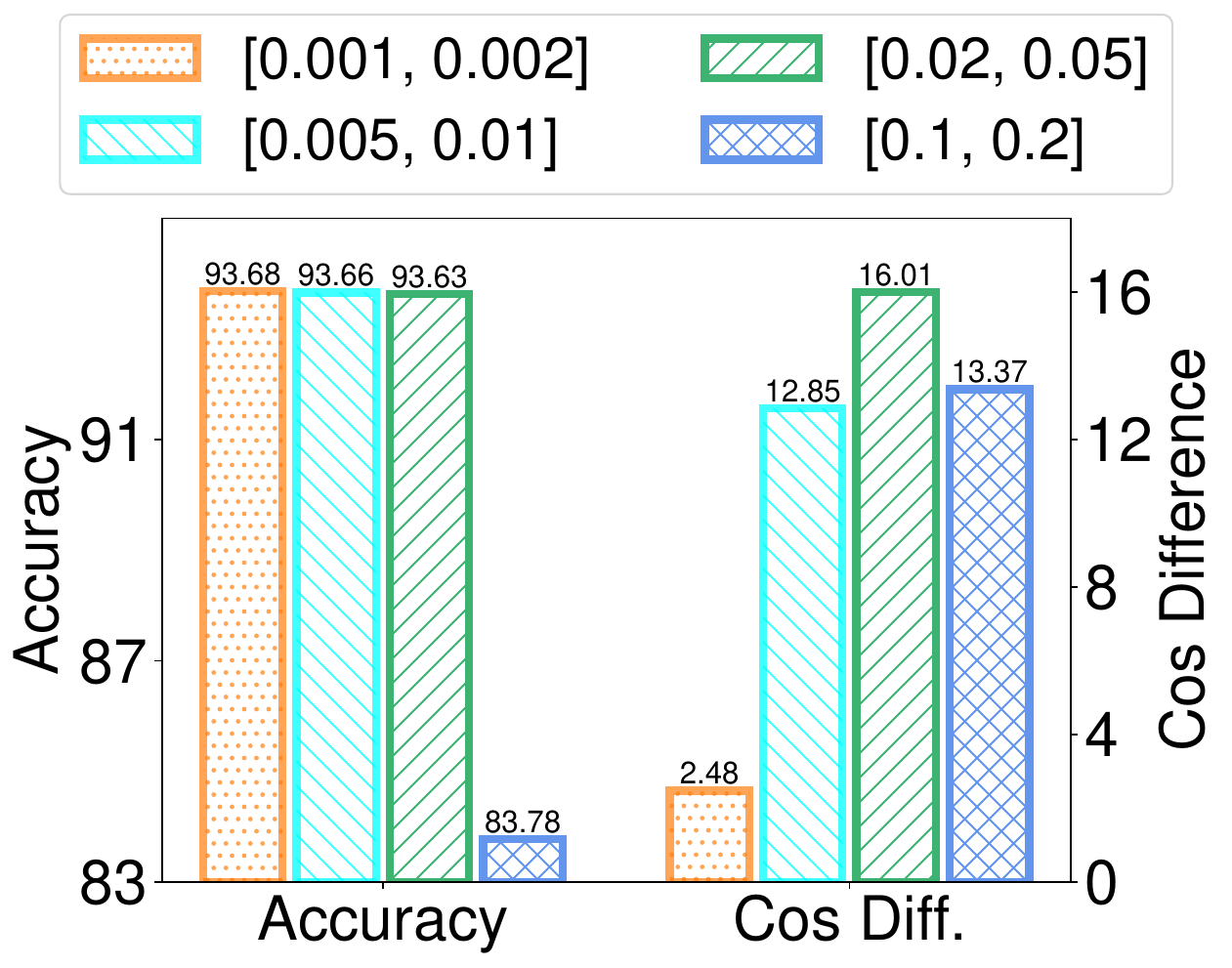}}
    \caption{The impact of the trigger set size $n$, the maximum number of triggers to fully activate watermark $m$, and the frequency interval on the AG News dataset.}
    \label{fig:hyper-ag}
\end{figure*}

\begin{figure*}[!t]
    \centering
    \subfigure[trigger set size $n$]{\label{fig:sel-enron}\includegraphics[width=0.32\textwidth]{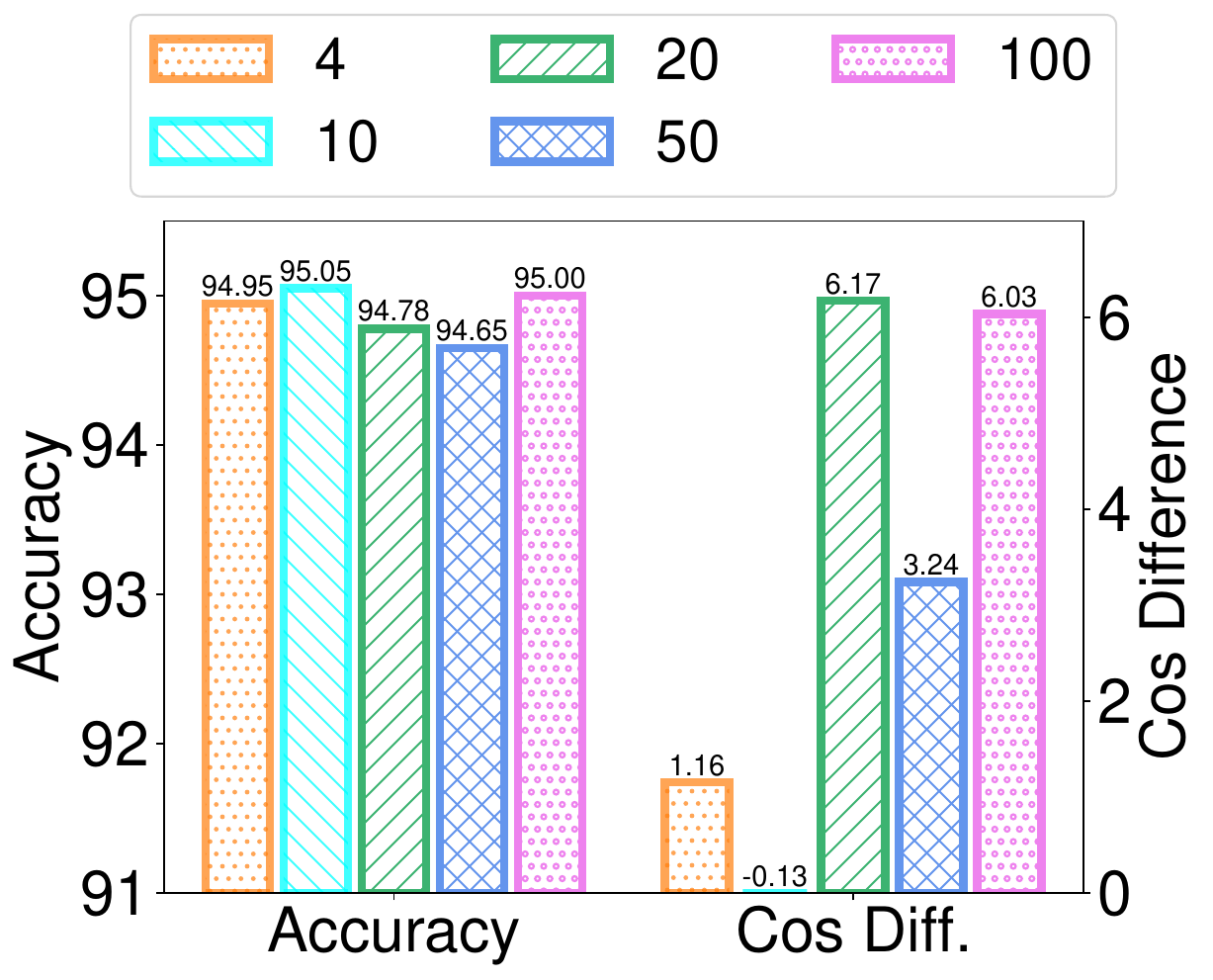}}
    \subfigure[max trigger number $m$]{\label{fig:max-enron}\includegraphics[width=0.32\textwidth]{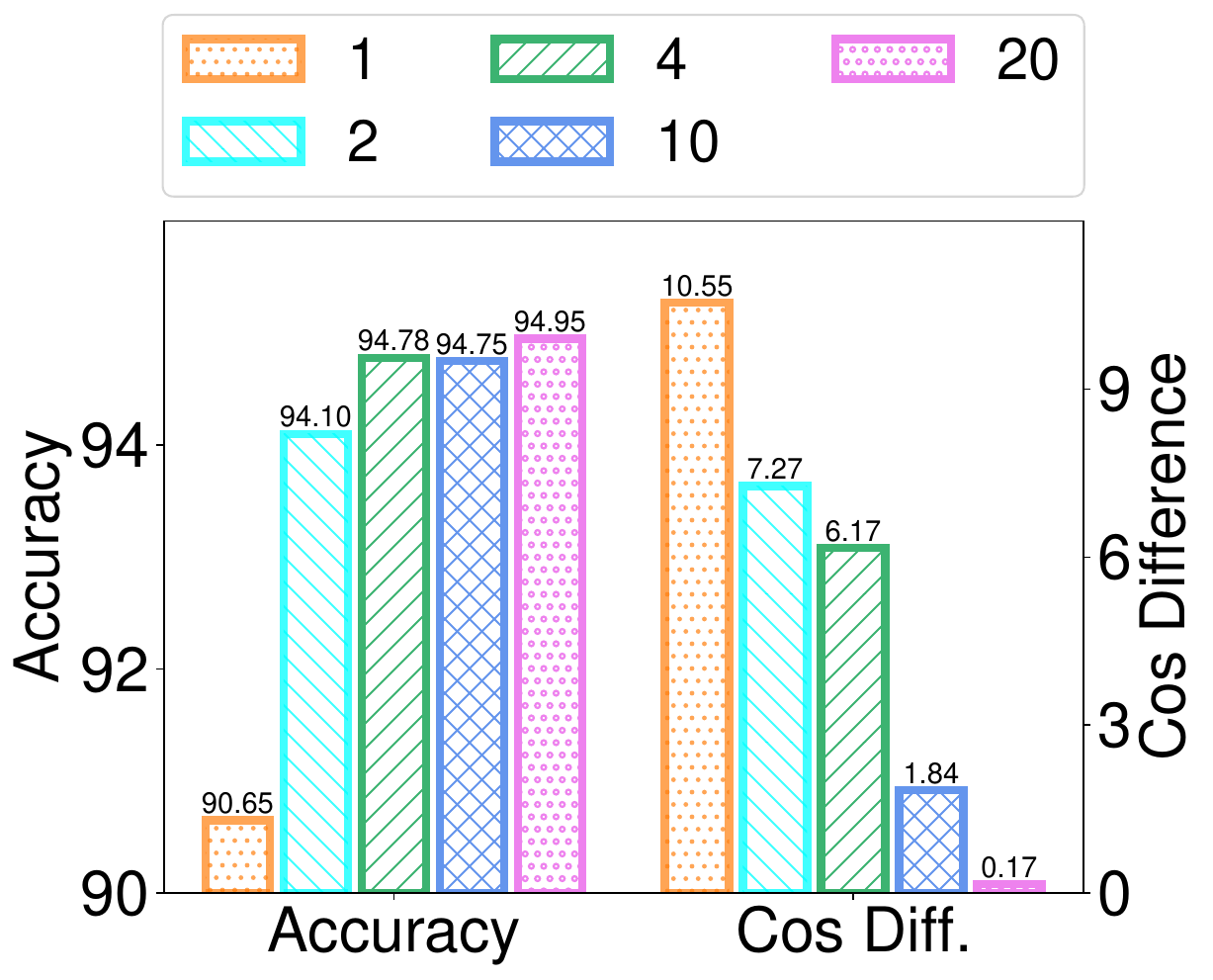}}
    \subfigure[frequency interval]{\label{fig:freq-enron}\includegraphics[width=0.32\textwidth]{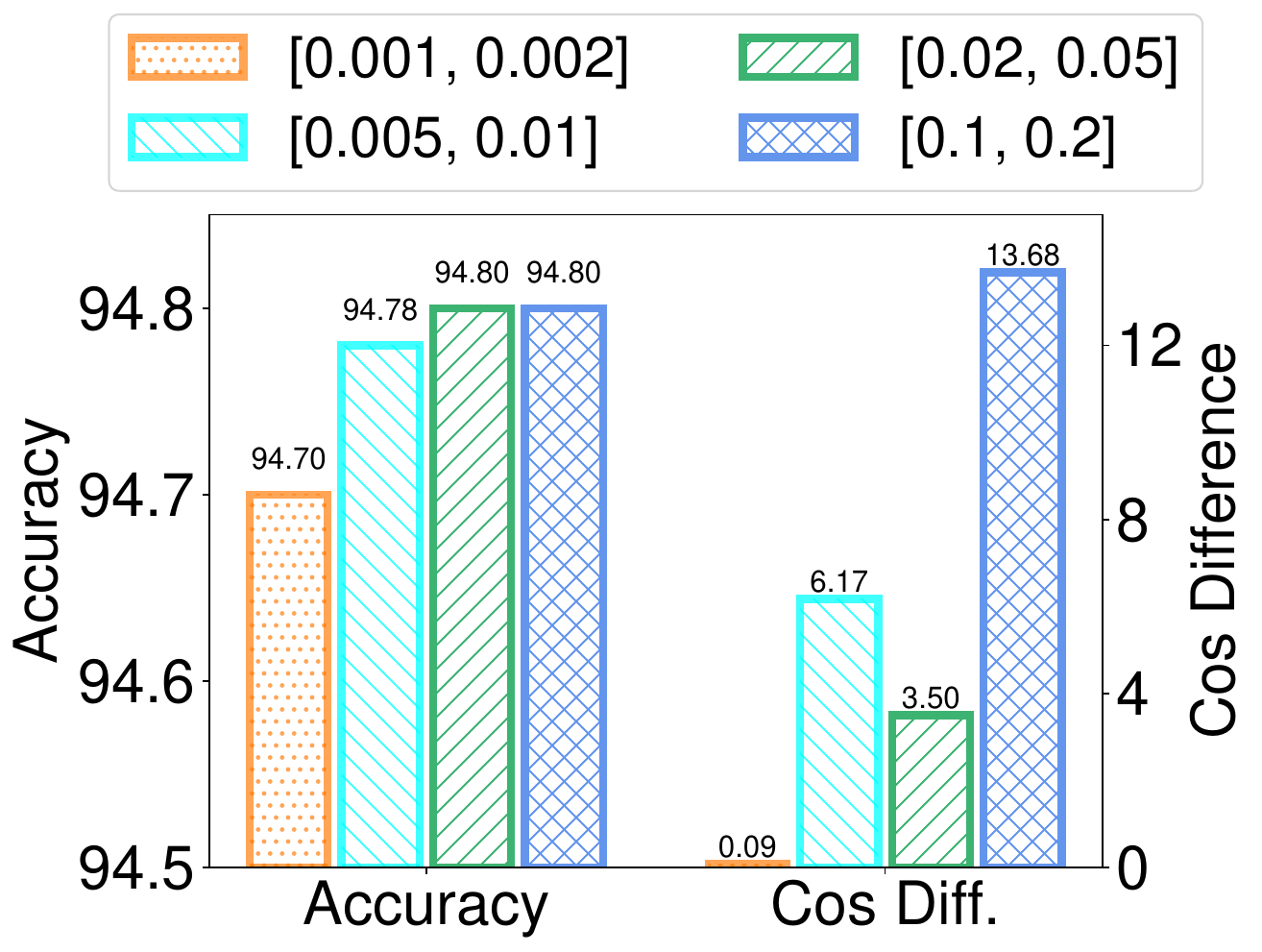}}
    \caption{The impact of the trigger set size $n$, the maximum number of triggers to fully activate watermark $m$, and the frequency interval on the Enron Spam dataset.}
    \label{fig:hyper-enron}
\end{figure*}

In this section, we show the experimental results of hyper-parameter analysis on MIND, Enron Spam and AG News datasets in Figure~\ref{fig:hyper-mind}, Figure~\ref{fig:hyper-ag}, Figure~\ref{fig:hyper-enron}, respectively.
Since the results of the visualization of PCA and t-SNE are too large to display on the paper, we put them in our repository.
The observations are almost the same as those we described in Section~\ref{sec:hyper}.
First, too small trigger set $n$ leads to low detection performance.
This is because the number of backdoor samplers is small with too small sizes of trigger sets, which reduces the likelihood of the extracted model inheriting the watermark.
Second, the trigger set $n$ has little impact on accuracy.
It might be because the frequency interval $[0.005, 0.01]$ is small.
Though the trigger set is large, the probability of 4 triggers appearing in a sentence is still low.
Third, we find that small $m$, especially $1$, degrades accuracy, while large $m$ reduces detection performance.
This is because about 1\% embeddings equal the pre-defined target embedding $\textbf{e}_t$ with $m=1$, which negatively impacts the provided embedding effectiveness.
When $m$ is large, the backdoor degree of most samples is too small to make the watermark inherited by the extracted model.
Finally, low frequencies bring negative impacts on detection performance, and high frequencies might negatively affect accuracy.
This is because high frequencies poison many embeddings and affect the performance of the provided embeddings.
In low-frequency settings, the watermark is only added to a few samples, which limits the possibility of watermark inheritance. 
Additionally, we analyze the impact of dropout values on model extraction attacks. 
When the dropout value is greater than 0.4, the model cannot be extracted effectively, rendering the detection ability of \method meaningless.
Therefore, in Table~\ref{tab:dropout_sst2}, we present the performance of \method when the dropout value is between 0 and 0.4.
Our observations indicate that model extraction attacks are most effective when the dropout value was set to 0.
This is because the LLM embeddings contain rich semantic knowledge, and increasing the dropout value weakens the stealer's model fitting ability, thereby reducing its performance in downstream tasks and the likelihood of inheriting watermarks.

\begin{table}[H]
\centering
\scalebox{0.87}{
\begin{tabular}{cccc}
\Xhline{1.5pt}
\multirow{2}{*}{Dropout Value} & \multicolumn{3}{c}{Detection Performance} \\ \cline{2-4} 
     & p-value        & $\Delta_{cos} (\%)$        & $\Delta_{l2} (\%)$                 \\ \hline
0.0        & $<10^{-5}$           & 4.07        & -8.13              \\
0.2        & $<10^{-7}$            & 2.82       & -5.65            \\
0.4      & $<3 \times 10^{-4}$           & 0.87        & -2.59             \\ 
\Xhline{1.5pt}
\end{tabular}
}
\caption{The impact of the dropout value used in FFN network on SST2.}
\label{tab:dropout_sst2}
\end{table}
\section{Theoretical Proof}
In this section, we provide theoretical proof for proportions in Section~\ref{sec:attack}.

\subsection{Proof of Proportion 1}
\label{sec:ds-invariant}
\textit{Proof.}
Given any pair of vectors $(\textbf{i}, \textbf{j})$, according to the definition of identity transformation, we have 

\begin{equation}
\begin{aligned}
        &||\frac{\mathbf{I}(\textbf{i})}{||\mathbf{I}(\textbf{i})||} - \frac{\mathbf{I}(\textbf{j})||^2}{||\mathbf{I}(\textbf{j})||} = ||\frac{\textbf{i}}{||\textbf{i}||} - \frac{\textbf{j}}{||\textbf{j}||}||_2^2, \\
        &cos(\mathbf{I}(\textbf{i}), \mathbf{I}(\textbf{j})) = cos(\textbf{i}, \textbf{j}),
\end{aligned}
\nonumber
\end{equation}
which indicates identity transformation is similarity-invariant.

For dimension-shift transformation $\mathbf{S}$, we have
\begin{equation}
\begin{aligned}
    &||\frac{\mathbf{S}(\textbf{i})}{||\mathbf{S}(\textbf{i})||} -\frac{\mathbf{S}(\textbf{j})}{||\mathbf{S}(\textbf{j})||}||^2 \\
    &= \sum_{k=1}^{d} (\frac{i_k}{||\textbf{i}||} - \frac{j_k}{||\textbf{j}||})^2
    = ||\frac{\textbf{i}}{||\textbf{i}||} - \frac{\textbf{j}}{||\textbf{j}||}||^2,
\end{aligned}      
\nonumber
\end{equation}

\begin{equation}
\begin{aligned}
     cos(\mathbf{S}(\textbf{i}), \mathbf{S}(\textbf{j})) &= \frac{\sum_{k=1}^{d} i_k j_k}{||\textbf{i}|| \,||\textbf{j}||}
    =  cos(\textbf{i}, \textbf{j}),
\end{aligned}      
\nonumber
\end{equation}
where $d$ is the dimension of $\textbf{i}$ and $\textbf{j}$. Therefore, dimension-shift transformation $\mathbf{S}$ is similarity-invariant as well.

\subsection{Proof of Proportion 2}
\label{sec:defend-all-ds}
\textit{Proof.} Denote the embedding of copied model as $\textbf{e}$, the embedding manipulated by transformation $\mathbf{A}_1$ as $\textbf{e}^1$ and the the embedding manipulated by transformation $\mathbf{A}_2$ as $\textbf{e}^2$.
Since both $\mathbf{A}_1$ and $\mathbf{A}_2$ are similarity-invariant, we have 
\begin{equation}
    \begin{aligned}
    &cos_i^1 = cos_i^2 = cos_i = \frac{\textbf{e}_i \cdot \textbf{e}_t'}{||\textbf{e}_i||\, ||\textbf{e}_t'||}, \\
    &l_{2i}^1 = l_{2i}^2 = l_{2i} = ||\textbf{e}_i/||\textbf{e}_i|| - \textbf{e}_t'/||\textbf{e}_t'|| \,||^2,
    \end{aligned}
\nonumber
\end{equation}
where the superscript indicates the similarity calculated under which transformation.
Therefore, we can obtain:
\begin{equation}
    \begin{aligned}
    &C_b^1 = C_b^2, C_n^1 = C_n^2, L_b^1 = L_b^2, L_n^1 = L_n^2.
    \end{aligned}
\nonumber
\end{equation}

Since the inputs for the metrics $\Delta_{cos}$, $\Delta_{l2}$ and p-value in our methods are only $C_b$, $C_n$, $L_b$ and $L_n$, we have
\begin{equation}
    \Delta_{cos}^1 = \Delta_{cos}^2, \Delta_{l2}^1 = \Delta_{l2}^2, p_{KS}^1 = p_{KS}^2,
\nonumber
\end{equation}
where $p_{KS}$ is the p-value of the KS test with $C_b$ and $C_n$ as inputs.
\section{Experimental Environments}

We conduct experiments on a linux server with Ubuntu 18.04.
The server has a V100-16GB with CUDA 11.6.
We use pytorch 1.13.1.


\end{document}